\documentclass[letterpaper, 10 pt, conference]{ieeeconf}

\IEEEoverridecommandlockouts

\usepackage{cite}
\usepackage{comment}
\usepackage{amsmath, amssymb, amsfonts}
\usepackage{algorithm}
\usepackage{algpseudocode}
\usepackage{graphicx}
\usepackage{textcomp}
\usepackage{stfloats}
\usepackage{xcolor}
\usepackage{paralist}
\usepackage{caption}
\usepackage{subcaption}

\DeclareMathOperator*{\argmin}{arg\,min}

\newcommand{\fr}[1]{\textcolor{red}{[FR: #1]}}

\def\BibTeX{{\rm B\kern-.05em{\sc i\kern-.025em b}\kern-.08em
    T\kern-.1667em\lower.7ex\hbox{E}\kern-.125emX}}

\usepackage{acro}
\usepackage{bm}
\usepackage{mathtools}
\usepackage{xcolor}

\DeclareAcronym{APhI}{short = APhI, long = Aerial Physical Interaction}
\DeclareAcronym{AM}{short = AM, long = Aerial Manipulator}
\DeclareAcronym{AR}{short = AR, long = Augmented Reality}
\DeclareAcronym{API}{short = API, long = Application Programming Interface}
\DeclareAcronym{BW}{short = BW, long = Bandwidth}
\DeclareAcronym{BIBO}{short = BIBO, long = Bounded Input–Bounded Output}
\DeclareAcronym{CNMPC}{short = CNMPC, long = Centralized Nonlinear Model Predictive Control}
\DeclareAcronym{CBF}{short = CBF, long = Control Barrier Function}
\DeclareAcronym{COM}{short = CoM, long = Center of Mass}
\DeclareAcronym{CPU}{short = CPU, long = Central Processing Unit}
\DeclareAcronym{CRI}{short = CRI, long = Container Runtime Interface}
\DeclareAcronym{DoF}{short = DoF, long = degrees of freedom}
\DeclareAcronym{DL}{short = DL, long = Deep Learning}
\DeclareAcronym{DRL}{short = DRL, long = Deep Reinforcement Learning}
\DeclareAcronym{DCS}{short = DCS, long = Distributed Control System}
\DeclareAcronym{DDS}{short = DDS, long = Data Distribution Service}
\DeclareAcronym{FT}{short = F/T, long = force and torque, short-indefinite = an, long-indefinite = a}
\DeclareAcronym{FPGA}{short = FPGA, long = Field-Programmable Gate Array}
\DeclareAcronym{GPS}{short = GPS, long = Global Positioning System}
\DeclareAcronym{GNSS}{short = GNSS, long = Global Navigation Satellite System}
\DeclareAcronym{GPU}{short = GPU, long = Graphics Processing Unit}
\DeclareAcronym{GUI}{short = GUI, long = Graphical User Interface}
\DeclareAcronym{HRI}{short = HRI, long = Human-Robot Interaction}
\DeclareAcronym{HMI}{short = HMI, long = Human Machine Interface}
\DeclareAcronym{ISM}{short = ISM, long = Industrial-Scientific-Medical}
\DeclareAcronym{IP}{short = IP, long = Internet Protocol}
\DeclareAcronym{I/O}{short = I/O, long = Input/Output}
\DeclareAcronym{IMU}{short = IMU, long = Inertial Measurement Unit}
\DeclareAcronym{k8s}{short = k8s, long = Kubernetes}
\DeclareAcronym{LQR}{short = LQR, long = Linear–Quadratic Regulator}
\DeclareAcronym{LTI}{short = LTI, long = Linear Time-Invariant}
\DeclareAcronym{LIO}{short = LIO, long = LiDAR Inertial Odometry}
\DeclareAcronym{LiDaR}{short = LiDaR, long = Light Detection and Ranging}
\DeclareAcronym{LoS}{short = LoS, long = Line of Sight}
\DeclareAcronym{MAV}{short = MAV, long = Micro Aerial Vehicle}
\DeclareAcronym{MBE}{short = MBE, long = momentum-based wrench estimator}
\DeclareAcronym{MPC}{short = MPC, long = Model Predictive Control}
\DeclareAcronym{MR}{short = MR, long = Mixed Reality}
\DeclareAcronym{MoCap}{short = MoCap, long = Motion Capture System}
\DeclareAcronym{ML}{short = ML, long = Machine Learning}
\DeclareAcronym{MIMO}{short = MIMO, long = Multiple Input Multiple Output}
\DeclareAcronym{MISO}{short = MISO, long = Multiple Input Single Output}
\DeclareAcronym{NDT}{short = NDT, long = non-destructive testing}
\DeclareAcronym{NMPC}{short = NMPC, long = Nonlinear Model Predictive Control}
\DeclareAcronym{OMAV}{short = OMAV, long = Omnidirectional Micro Aerial Vehicle}
\DeclareAcronym{PEMS}{short = PEMS, long = Power and Energy Monitoring System}
\DeclareAcronym{PTC}{short = PTC, long = pose tracking controller}
\DeclareAcronym{PID}{short = PID, long = Proportional-Integral-Derivative}
\DeclareAcronym{PPO}{short = PPO, long = Proximal Policy Optimization}
\DeclareAcronym{PETG}{short = PETG, long = Polyethylene Terephthalate Glycol}
\DeclareAcronym{PoI}{short = PoI, long = Point-of-Interest}
\DeclareAcronym{PLC}{short = PLC, long = Programmable Logic Controller}
\DeclareAcronym{PCS}{short = PCS, long = Process Control System}
\DeclareAcronym{QoS}{short = QoS, long = Quality of Service}
\DeclareAcronym{RAM}{short = RAM, long = Random Access Memory}
\DeclareAcronym{RGB}{short = RGB, long = Red Green Blue}
\DeclareAcronym{RGB-D}{short = RGB-D, long = Red Green Blue - Depth}
\DeclareAcronym{RL}{short = RL, long = Reinforcement Learning}
\DeclareAcronym{RRT}{short = RRT, long = round-trip-time}
\DeclareAcronym{ROS}{short = ROS, long = Robot Operating System}
\DeclareAcronym{RBAC}{short = RBAC, long = Role-Based Access Control}
\DeclareAcronym{SLAM}{short = SLAM, long = Simultaneous Localization And Mapping}
\DeclareAcronym{SNR}{short = SNR, long = signal-to-noise ratio}
\DeclareAcronym{SINR}{short = SINR, long = signal-to-interference-plus-noise ratio}
\DeclareAcronym{SWR}{short = SWR, long = Standing Wave Ratio}
\DeclareAcronym{SCADA}{short = SCADA, long = Supervisory Control and Data Acquisition}
\DeclareAcronym{SISO}{short = SISO, long = Single Input Single Output}
\DeclareAcronym{SIMO}{short = SIMO, long = Single Input Multiple Output}
\DeclareAcronym{TCP}{short = TCP, long = Transmission Control Protocol}
\DeclareAcronym{TPU}{short = TPU, long = Thermoplastic Polyurethanes}
\DeclareAcronym{UAV}{short = UAV, long = Unmanned Aerial Vehicle}
\DeclareAcronym{UDP}{short = UDP, long = User Datagram Protocol}
\DeclareAcronym{UHF}{short = UHF, long = Ultra High Frequency (300 MHz - 1300 MHz)}
\DeclareAcronym{VR}{short = VR, long = Virtual Reality}
\DeclareAcronym{VM}{short = VM, long = Virtual Machine}
\DeclareAcronym{VPN}{short = VPN, long = Virtual Private Network}
\DeclareAcronym{VHF}{short = VHF, long = Very High Frequency (30 MHz - 300 MHz)}
\DeclareAcronym{VIO}{short = VIO, long = Visual Inertial Odometry}
\DeclareAcronym{WTC}{short = WTC, long = wrench tracking controller}
\DeclareAcronym{YOLO}{short = YOLO, long = You Only Look Once}
\DeclareAcronym{YLLO}{short = YLLO, long = You Look Less than Once}

\renewcommand{\vec}[1]{\bm{#1}}		
\newcommand{\nR}[1]{\mathbb{R}^{#1}}		
\newcommand{\upperRomannumeral}[1]{\uppercase\expandafter{\romannumeral#1}}	

\renewcommand{\frame}[1]{\mathcal{F}_{#1}}		

\newcommand{\posA}{\vec{p}_A}				
\newcommand{\velA}{\vec{v}_A}				
\newcommand{\accA}{\dot{\vec{v}}_A^A}				

\newcommand{\frameB}{\frame{B}}			
\newcommand{\frameA}{\frame{A}}			
\newcommand{\frameC}{\frame{C}}			
\newcommand{\frameT}{\frame{T}}			

\begin{document}
\title{Assisted Physical Interaction: Autonomous Aerial Robots with \\ Neural Network Detection, Navigation, and Safety Layers}
\author{Andrea Berra$^{*1}$, Viswa Narayanan Sankaranarayanan$^{*2}$, Achilleas Santi Seisa$^{*2}$, Julien Mellet$^{*3}$, Udayanga \\G.W.K.N. Gamage$^{*4}$, Sumeet Gajanan Satpute$^{2}$, Fabio Ruggiero$^{3}$, Vincenzo Lippiello$^{3}$, Silvia Tolu$^{4}$, \\Matteo Fumagalli$^{4}$, George Nikolakopoulos$^{2}$, Miguel Ángel Trujillo Soto$^{1}$, and Guillermo Heredia$^{5}$
\thanks{This project has received funding from the European Union’s Horizon 2020 research and innovation program under the Marie Skłodowska-Curie grant agreement No 953454.}
\thanks{$^{1}$CATEC, Advanced Center for Aerospace Technologies, Seville, Spain.}
\thanks{$^{2}$Robotics and AI Team, Department of Computer, Electrical and Space Engineering, Lule\aa University of Technology, Lule\aa, Sweden.}
\thanks{$^{3}$PRISMA Lab, Department of Electrical Engineering and Information Technology, University of Naples Federico II Naples, Italy.}
\thanks{$^{4}$Automation and Control Group, Department of Electrical Engineering and Photonics, Technical University of Denmark, Denmark.}
\thanks{$^{5}$Robotics, Vision, and Control Group School of Engineering, University of Seville, Seville, Spain.}
\thanks{$^{*}$The authors contributed equally}
\thanks{Corresponding Authors' email: {\tt\small \{vissan, achsei\}@ltu.se, aberra@catec.aero, julien.mellet@unina.it, kniud@dtu.dk}}
}

\maketitle

\begin{abstract}
The paper introduces a novel framework for safe and autonomous aerial physical interaction in industrial settings. It comprises two main components: a neural network-based target detection system enhanced with edge computing for reduced onboard computational load, and a control barrier function (CBF)-based controller for safe and precise maneuvering.
The target detection system is trained on a dataset under challenging visual conditions and evaluated for accuracy across various unseen data with changing lighting conditions. Depth features are utilized for target pose estimation, with the entire detection framework offloaded into low-latency edge computing.
The CBF-based controller enables the UAV to converge safely to the target for precise contact. Simulated evaluations of both the controller and target detection are presented, alongside an analysis of real-world detection performance.
\end{abstract}

\begin{keywords}
Aerial Physical Interaction; UAVs; Control Barrier Function; Neural Network; Edge Computing.
\end{keywords}

\section{Introduction} 
\label{sect:intro}

\begin{figure}[t!]
    \centering
    \includegraphics[width=0.96\columnwidth]{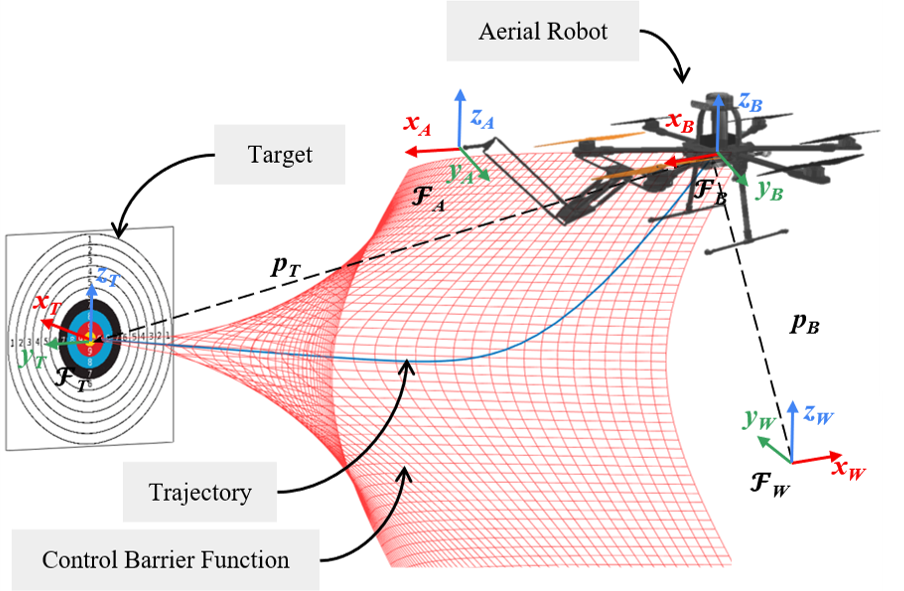}
    \caption{Schematic of the boundary of the control barrier function, which guides the aerial platform's trajectory to reach the target precisely.}
    \label{fig:frames}
    \vspace{-0.5cm}
\end{figure}

Operating \ac{MAV}s for inspections and maintenance is crucial in industries, as demonstrated in projects such as ARCAS and AEROARMS~\cite{ollero2019aerial}. 
It is particularly pertinent in hazardous and hard-to-reach environments.
Autonomous \ac{UAV}s have improved reliability and safety, especially in the absence of \ac{LoS} \cite{nikolakopoulos2022aerial}. Subsequently, aerial manipulation has become a rapidly growing area of research, offering interesting challenges in the \ac{PoI} localization and ensuring safe navigation, while incorporating diverse robotic systems for autonomous \ac{APhI} with the environment~\cite{past-future-am}. Since \ac{UAV}s with their razor-sharp propellers can wreak havoc during a crash, it is essential to ensure security at different levels.
Conventional methods often entail the use of an expert operator~\cite{bridge-inspection}, necessitating complete supervision combined with high mental demand. On the other hand, autonomous vehicles running fully onboard often lack safety features and are prompted by unexpected technical limitations.

Motivated by the consistent need for advanced safety for autonomous \ac{APhI}, this work aims to develop a pipeline for \ac{PoI} localization and establish precise and safe contact with it. Further, we seek to enhance the performance using a neural-based target detection strategy, coupled with edge computing.
To validate the proposed methodology, a realistic simulation incorporating an aerial manipulator and its surrounding environment is proposed. The simulated \ac{UAV} encompasses the aerial manipulator platform developed for experimental phases, coupled with the sensor setup and diverse simulated world environments. The inclusion of different world environments allows us to robustly evaluate the proposed approach under different environmental conditions.

\subsection{Related Work} 
\label{sect:related_work}

Neural networks have made significant advancements in target detection, especially in challenging \ac{UAV} image processing environments. Applications include precision agriculture for fruit classification~\cite{zhu2022rapid}, maritime surveillance~\cite{drones6110335}, forest fire detection~\cite{f14091812}, and search and rescue operations~\cite{s16111778}. The widely used \ac{YOLO} network~\cite{jiang2022review} is prevalent in aerial target detection due to its reliable performance and improved accuracy, reaching centimeter-level precision in \ac{UAV} applications~\cite{kou2020research}. Tailored adaptations of \ac{YOLO} for \ac{UAV} imagery conditions, such as occlusion and brightness variations, have also been proposed~\cite{TAN2021107261}.

However, many existing approaches require substantial computational resources beyond what is available on \ac{UAV} onboard computers. Edge computing has emerged as a crucial solution for computational offloading, offering superior computational power compared to onboard systems and reduced latency relative to cloud computing. For example, in~\cite{9372841}, a lightweight object recognition algorithm based on \ac{YLLO} runs on the edge for real-time traffic area monitoring with minimal data to preserve accuracy. Similarly, in~\cite{8479091}, edge computing supports real-time object identification and classification by multiple aerial and mobile robots using a \ac{YOLO} algorithm.

Safety and precision in aerial robotic systems have been approached from a Prescribed Performance Control (PPC) perspective~\cite{ganguly2021efficient, sankaranarayanan2022robustifying, sankaranarayanan2023adaptive}. While PPC ensures safety, it typically requires an external planner. However, recent advancements in \ac{CBF} offer an elegant solution for providing safety in dynamic systems without the need for an external planner~\cite{ames2019control}. \ac{CBF}s linearize constraints over the control space, making it suitable for navigation applications to run directly on the onboard computer.

In this context, while neural networks have shown promise for end-to-end control~\cite{SAVIOLO202345}, our approach focuses on leveraging model-based control for safety, providing a continuous definition of the output controller with a dynamic model~\cite{aerial_robotics_springer}. Although neural-based techniques excel in image processing tasks~\cite{o2020deep}, we opt to use them solely for detection purposes in our autonomous system. We complement this approach by offloading to edge computing platforms. While \ac{CBF}s have been utilized in aerial robots for collision avoidance~\cite{qing2021collision} and landing~\cite{lee2016vision}, their application for actively enforcing safe navigation to establish physical contact is novel and unexplored. While \cite{lippi2021control, ferraguti2020control} have proposed CBF approaches for physical interaction using a manipulator, they have been used in the context of avoidance. Similarly, \cite{liang2023adaptive} proposes a CBF approach for aerial interaction to improve the tracking efficiency for uncertainties. When it comes to \ac{APhI}, the need is to precisely enable contact at a given point, while avoiding collision with the target. The CBF used in this approach guides the \ac{UAV} to achieve this without an external planner.

\subsection{Contributions} 
\label{sect:contributions}
Given the identified shortcomings, our contributions encompass
\begin{inparaenum}
    \item the design of a dataset for computer vision adapted to the uncertainties of industrial environments;
    \item including edge computing for data processing;
    \item ensuring safe navigation and precise contact using a novel \ac{CBF}; and
    \item qualitative evaluation of the proposed system in some demanding conditions.
\end{inparaenum}

\section{Methodology} 
\label{sect:method}
\begin{figure}[t!]
    \centering
    \includegraphics[width=\columnwidth]{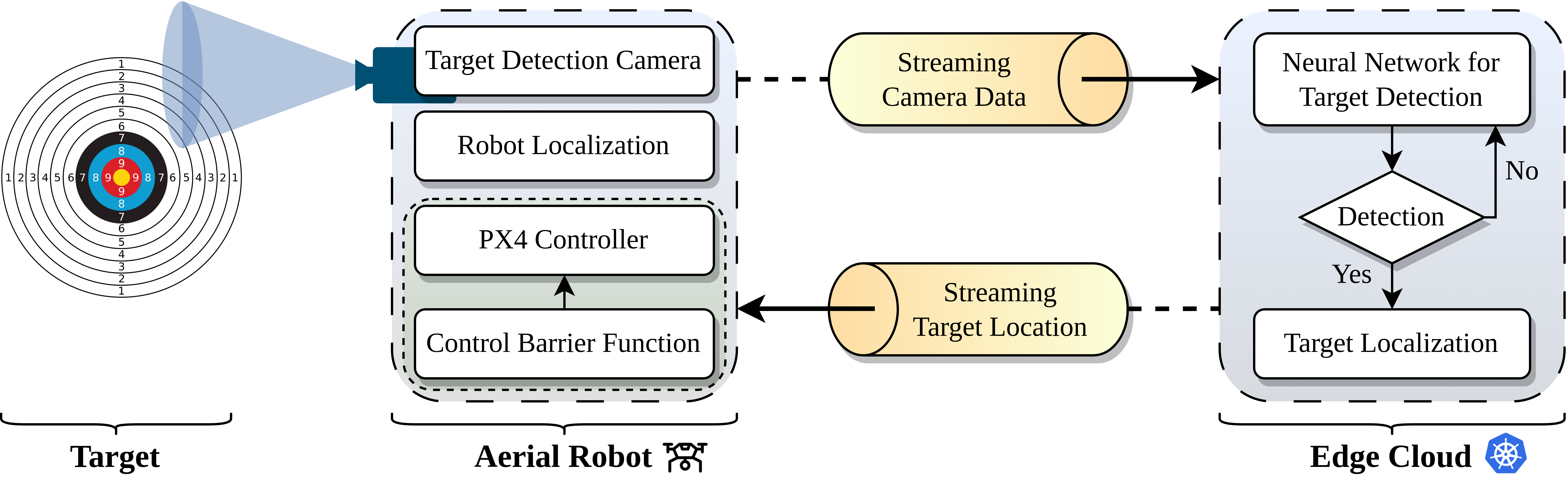}
    \caption{Schematic of the architecture incorporating edge offloading for the target detection and localization.}
    \label{fig:offloading}
    \vspace{-0.5cm}
\end{figure}

Our proposed pipeline comprises three main components: 
robot localization, \ac{PoI} or target localization, and safe navigation. The robot localization and navigation modules run onboard, while the computationally heavy target localization runs offboard on an edge computer (see Fig. \ref{fig:offloading}). Additionally, a simulation environment is developed to replicate the real setup for testing and remote visualization purposes. 
The following subsections provide a detailed discussion of each component. 

\subsection{Robot Localization} 
\label{sect:robot_loc}
A crucial aspect of autonomous navigation is accurately localizing the platform~\cite{lin2023gnss, 10275007}, which involves determining its pose relative to a fixed frame. However, traditional methods relying solely on \ac{GNSS} can be unreliable in metallic or underground environments. Hence, alternative methods such as \ac{VIO} become indispensable.

\begin{figure}[ht!]
    \centering
    \includegraphics[width=0.9\columnwidth]{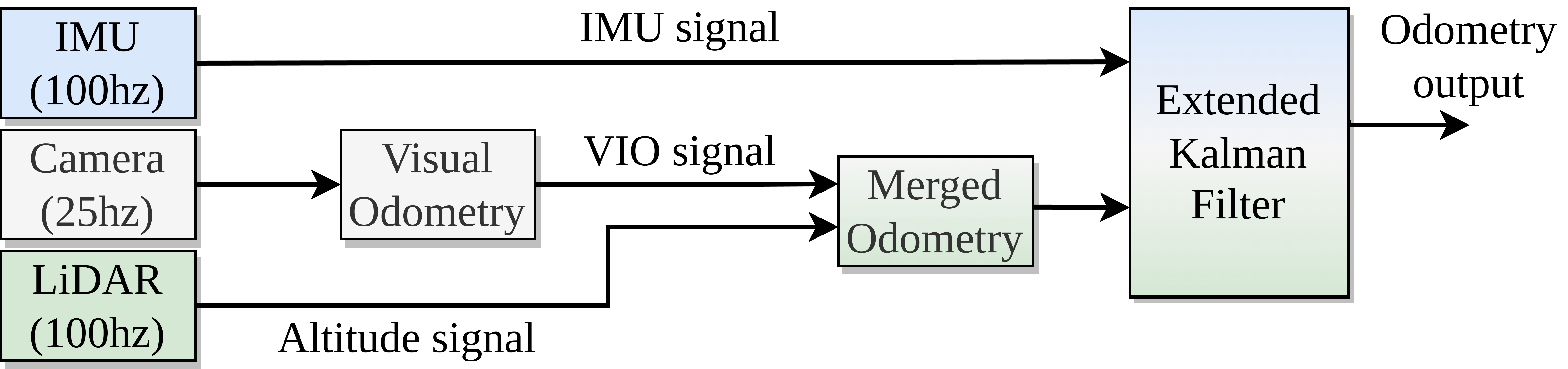}
    \caption{Block diagram illustrating the \ac{VIO} localization process, incorporating Visual Odometry, a micro \ac{LiDaR} sensor, and inertial measurements from the vehicle's \ac{IMU}.}
    \label{fig:odometry}
\end{figure}

In this work, we use RGBD-camera-based \ac{VIO} with the vehicle's onboard \ac{IMU} for estimating the \ac{UAV}'s pose~\cite{unknown-author-no-date}. We further improve the inaccuracies on altitude measurement using a single point micro \ac{LiDaR} (see Fig. \ref{fig:odometry}).

In addition, it is essential to determine both the pose of the manipulator's end-effector and the target's pose, defined by the frames $\frameA$ and $\frameT$, respectively (see Fig.~\ref{fig:frames}), as well as the pose of two cameras (odometry and target detection camera), relative to the \ac{UAV}'s body frame, referred to as $\frameB$. Hence, all the necessary transformations should be considered. This is crucial for enabling autonomous navigation and precise targeting of the object.

\subsection{Dataset Design} 
\label{sect:detection}
Fiducial marker detection and localization is a solved problem~\cite{5184844} in a controlled environment without occlusion. Nevertheless, the detection of a custom target with a \ac{RGB} camera is still challenging. Variations in lighting conditions, background clutter, occlusions, and other factors make traditional algorithms unreliable for safe deployment. While traditional computer vision techniques can lack adaptability~\cite{o2020deep}, neural networks are capable of learning from diverse data, making them robust to changes. We present our strategy to make a dataset ensuring safe target detection in an industrial context, that is later used to train a convolutional network.

\begin{figure}[t!] 
    \centering
    \begin{subfigure}{0.4\linewidth}
        \centering
        \includegraphics[width=\linewidth]{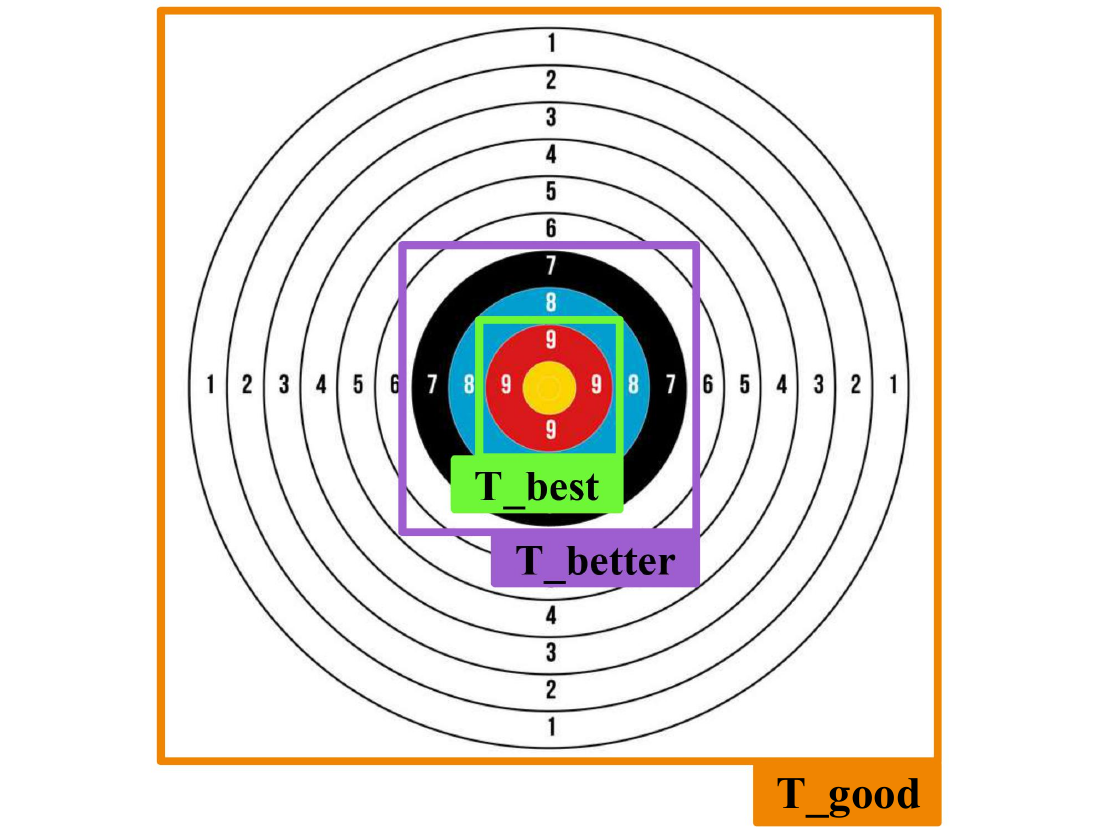}
        \caption{Labels associated to target detection levels}
        \label{fig:label-1}
    \end{subfigure}
    \hfill
    \begin{subfigure}{0.49\linewidth}
        \centering
        \includegraphics[width=\linewidth]{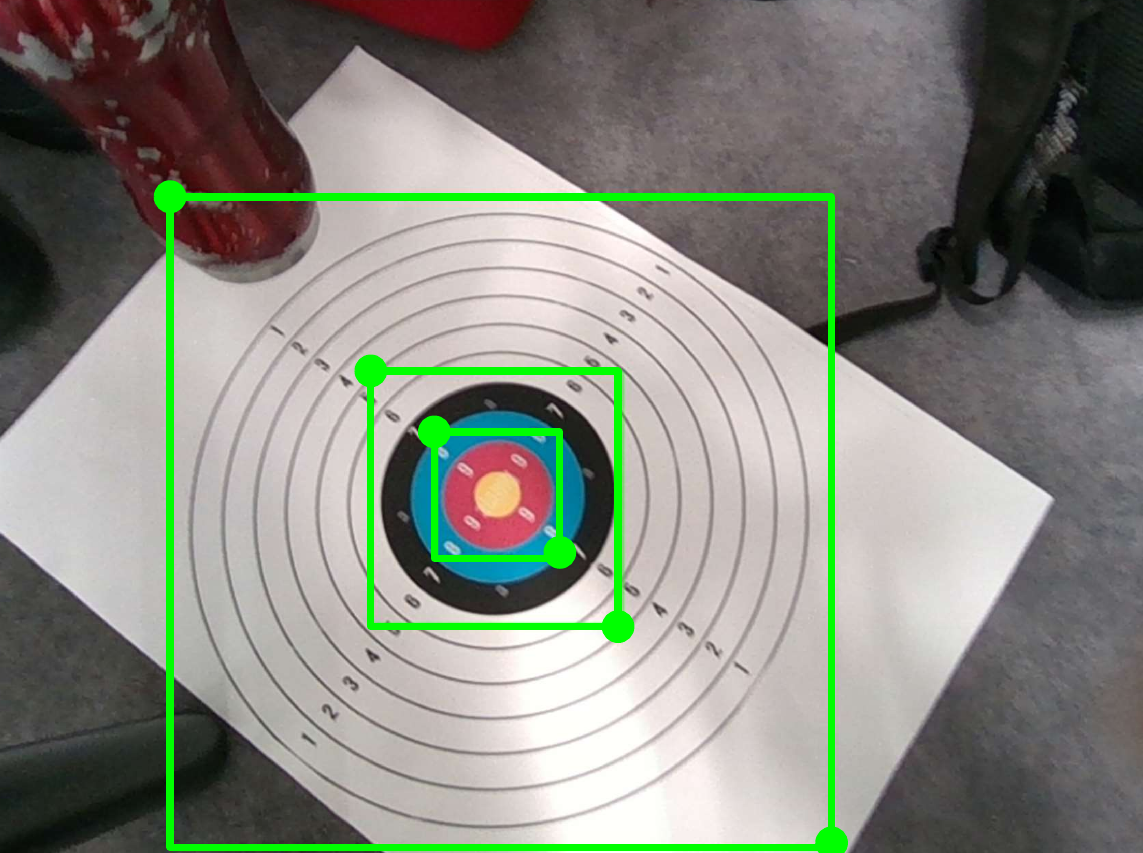}
        \caption{Labeling process from $labelme$ software~\cite{kentaro_wada_2021_5711226}}
        \label{fig:label-2}
    \end{subfigure}
    \caption{Image labels for target detection.}
    \label{fig:labels}
    \vspace{-0.5cm}
\end{figure}

We first define different levels of interest for the target (see Fig \ref{fig:labels}), allowing accurate target detection at different distances. In our case, we defined three levels:
\begin{inparaenum}
    \item $T\_good$ is the largest target circumscribing the following two,
    \item $T\_better$ is the intermediate one, surrounding the black circle, and
    \item $T\_best$ is the most accurate feature of the target, including only the two circles that make up the center of the target.
\end{inparaenum}
We labeled the training set, as presented in Fig.~\ref{fig:label-1}, while Fig.~\ref{fig:label-2} presents the view from $labelme$ software~\cite{kentaro_wada_2021_5711226}.

\begin{figure*}[b!] 
    \vspace{-0.5cm}
    \centering
    \begin{subfigure}{0.115\linewidth}
        \centering
        \includegraphics[width=\linewidth]{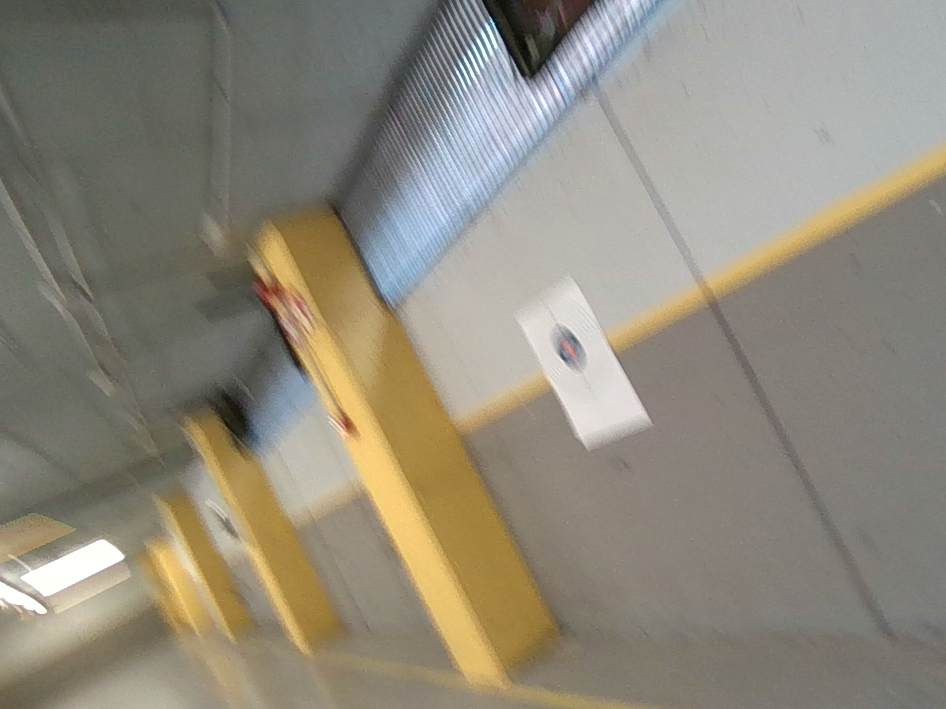}
        \caption{Blur}
        \label{fig:detection-1}
    \end{subfigure}
    \hfill
    \begin{subfigure}{0.115\linewidth}
        \centering
        \includegraphics[width=\linewidth]{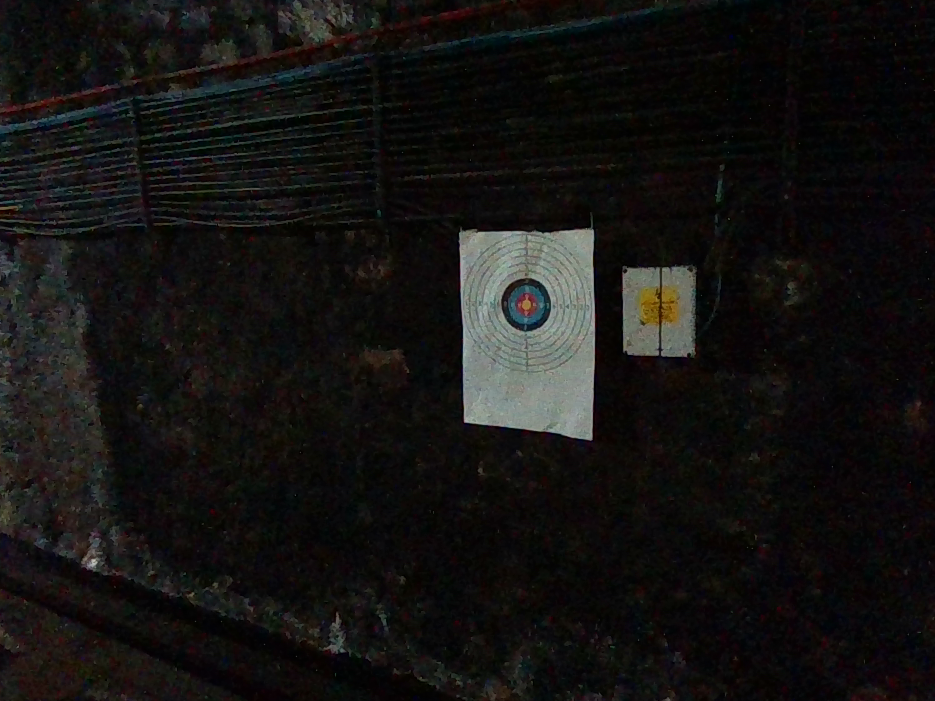}
        \caption{Low light}
        \label{fig:detection-2}
    \end{subfigure}
    \hfill
    \begin{subfigure}{0.12\linewidth}
        \centering
        \includegraphics[width=\linewidth]{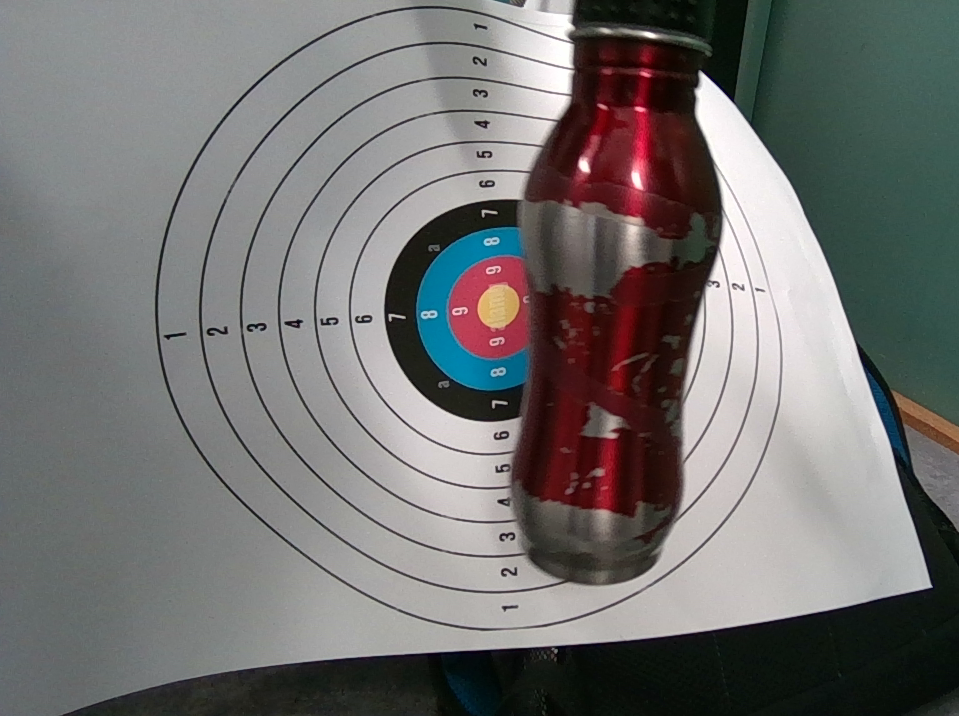}
        \caption{Obstruction}
        \label{fig:detection-3}
    \end{subfigure}
    \hfill
    \begin{subfigure}{0.115\linewidth}
        \centering
        \includegraphics[width=\linewidth]{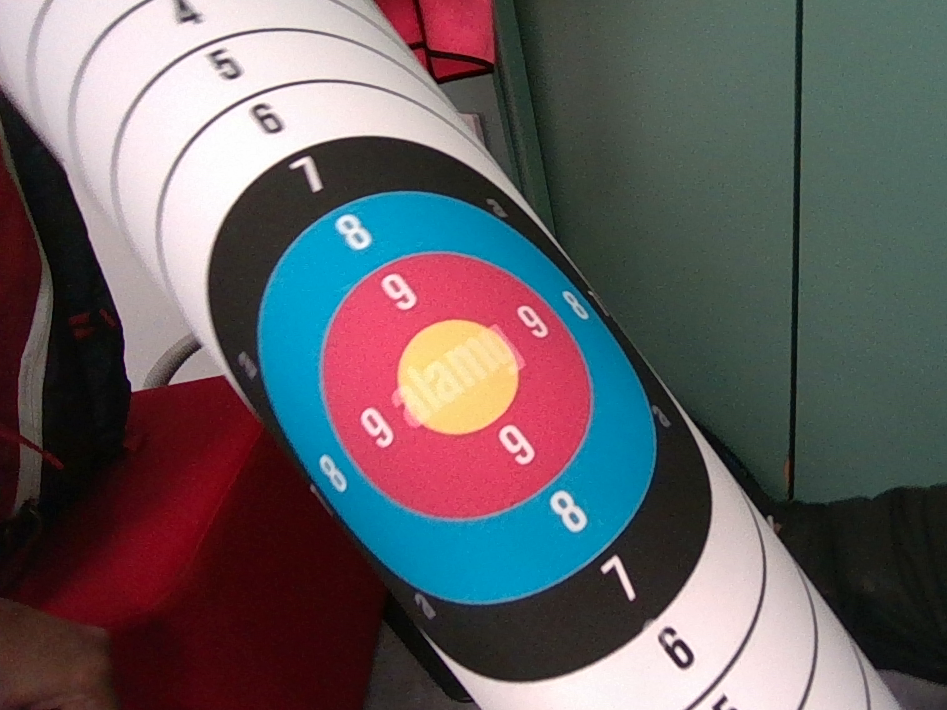}
        \caption{Distortion}
        \label{fig:detection-4}
    \end{subfigure}
    \hfill
    \begin{subfigure}{0.115\linewidth}
        \centering
        \includegraphics[width=\linewidth]{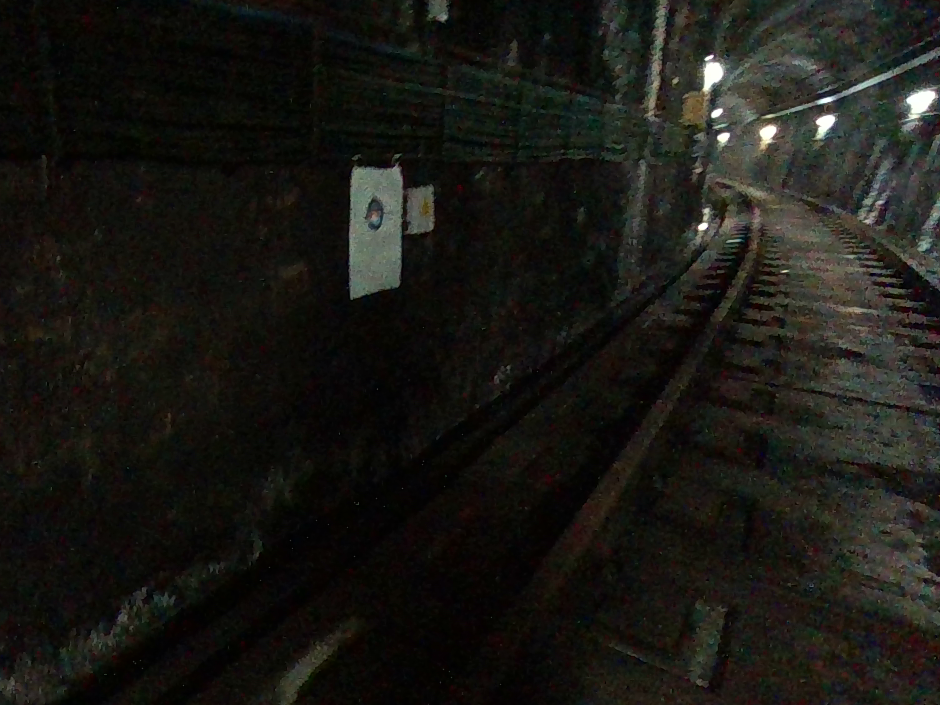}
        \caption{Far}
        \label{fig:detection-5}
    \end{subfigure}
    \hfill
    \begin{subfigure}{0.115\linewidth}
        \centering
        \includegraphics[width=\linewidth]{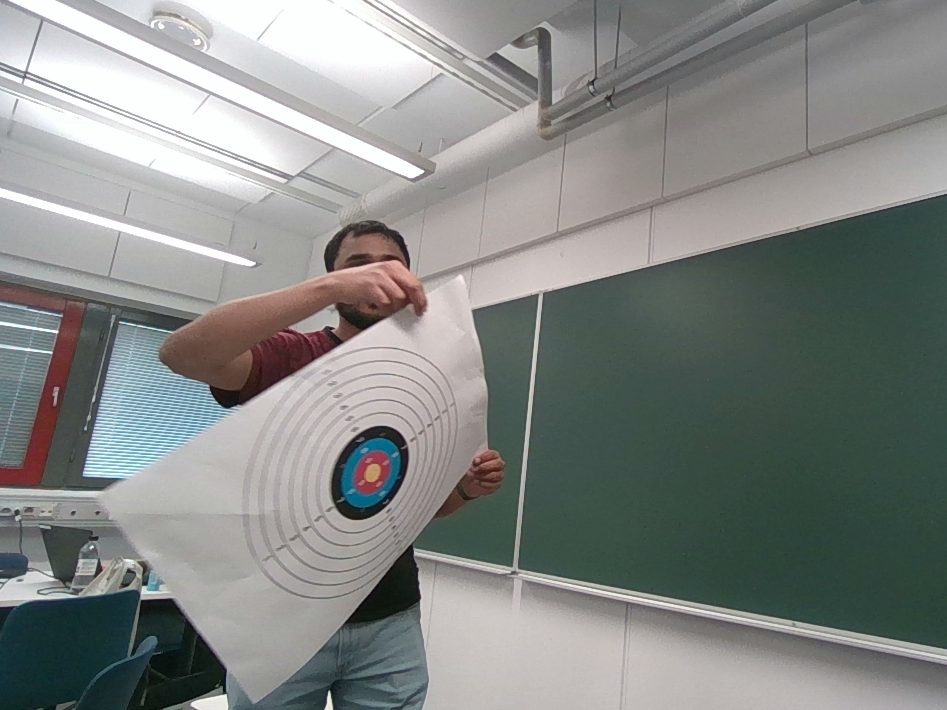}
        \caption{Orientation}
        \label{fig:detection-8}
    \end{subfigure}
    \hfill
    \begin{subfigure}{0.115\linewidth}
        \centering
        \includegraphics[width=\linewidth]{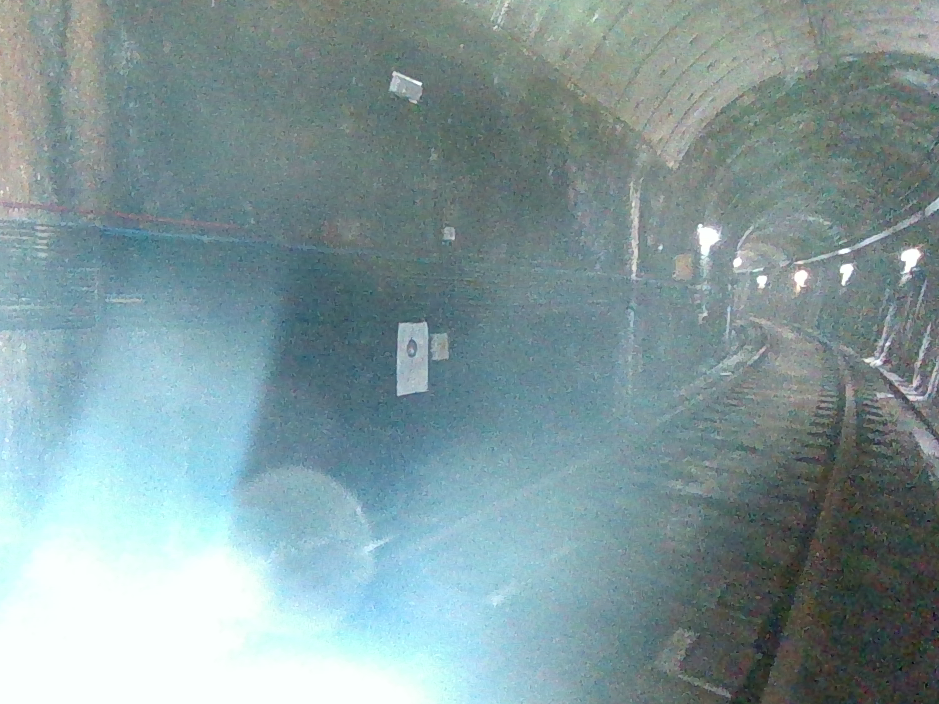}
        \caption{Flare}
        \label{fig:detection-9}
    \end{subfigure}
    \hfill
    \begin{subfigure}{0.115\linewidth}
        \centering
        \includegraphics[width=\linewidth]{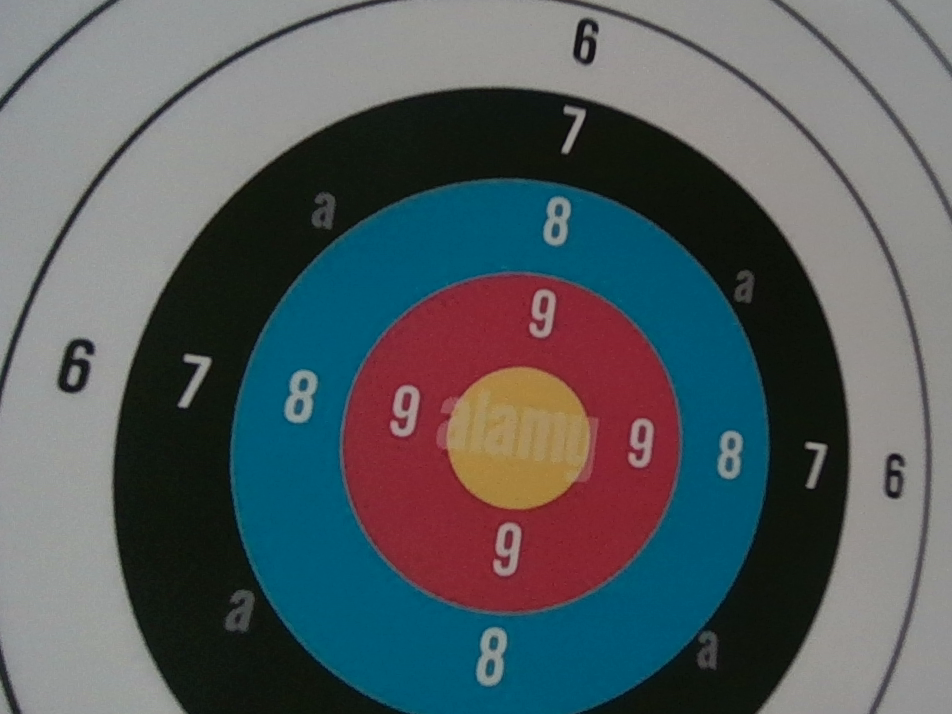}
        \caption{Close}
        \label{fig:detection-10}
    \end{subfigure}
    \caption{Samples from the dataset, compiling the environmental variables introduced to add noise.}
    \label{fig:cv-variations}
\end{figure*}

We prepared a dataset of $322$ labeled images. We first extracted $18$ images from an early flight test in a controlled laboratory environment. We increased the noise in the dataset with $146$ images providing obstruction, target surface distortion, various target orientations, and short distance (Figs.~\ref{fig:detection-3},~\ref{fig:detection-4},~\ref{fig:detection-8},~\ref{fig:detection-10}). Then, to increase the robustness of the target detection in specific demanding conditions, we fine-tuned the network using on-site data. We thus included $43$ images taken from a parking lot, adding blur and long-distance images (Figs.~\ref{fig:detection-1},~\ref{fig:detection-5}). We also included $20$ images taken on a bright day. The last demanding environment we obtained images from is a tunnel, with $48$ images in low-light conditions and flares on the camera (Figs.~\ref{fig:detection-2},~\ref{fig:detection-9}). We completed the dataset with $47$ images from the simulator. We thus propose a rich dataset with various environmental conditions corresponding to the specific industrial contexts.

We trained the \ac{YOLO}~\cite{jiang2022review} object detector using the aforementioned dataset. The detected target information is passed to the Target Localization as a bounding box.

\subsection{Target Localization} 
\label{sect:target_loc}
\begin{figure}[h!]
    \centering
    \includegraphics[width=0.5\textwidth]{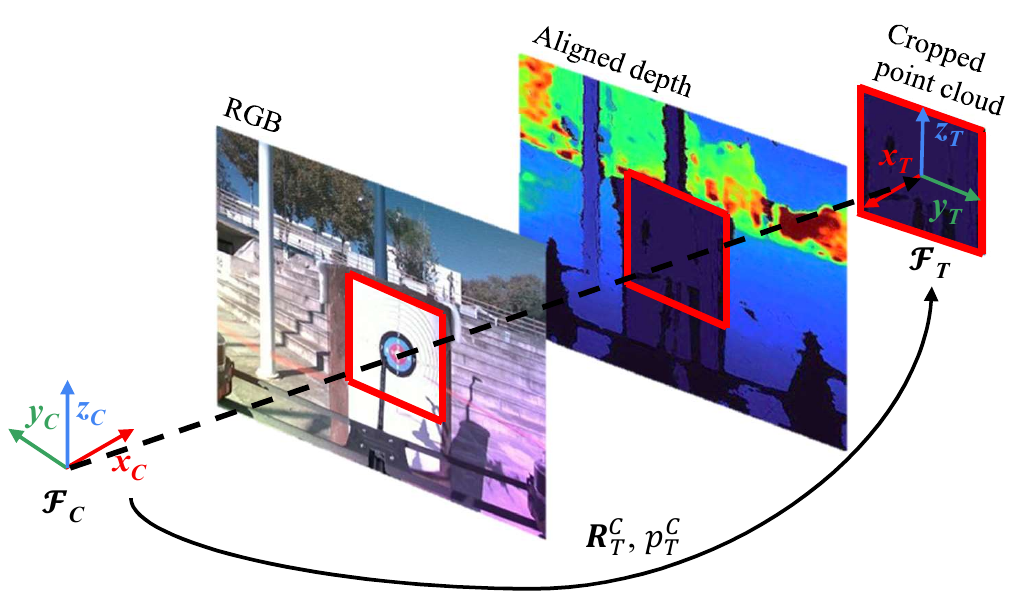}
    \caption{Estimation of position $p_T^C$ and orientation $\mathbf{R}_T^C$ of the target frame described by the cropped point cloud.}
    \label{fig:target_localization_approach}
    \vspace{-0.5cm}
\end{figure}

Target localization consists of estimating the target object pose from the camera frame using information both from the depth sensor and the color camera. The presented approach involves the estimation of the rotation and translation matrix that solves the difference between the 3D points of the point cloud and the 2D points projected into the camera frame $\frameC$, as illustrated in Fig. \ref{fig:target_localization_approach}. The solution to the problem, presented here, follows the Perspective-n-Points (PnP) approach introduced in~\cite{marchand2015pose},
\begin{equation}
\begin{bmatrix}
    u \\
    v \\
    1
\end{bmatrix}
    =
    \begin{bmatrix}
    f_x & 0 & c_x \\
    0 & f_y & c_y \\
    0 & 0 & 1
    \end{bmatrix}
    \begin{bmatrix}
    1 & 0 & 0 & 0 \\
    0 & 1 & 0 & 0 \\
    0 & 0 & 0 & 1
    \end{bmatrix}
    \begin{bmatrix}
        \mathbf{R}_T^C & p_T^C \\
        0 & 1
    \end{bmatrix}
    \Tilde{p}^T_T
    \label{eq:pnp_maths}
\end{equation}
where $p_T^C \in \nR{3}$ and $R_T^C \in \nR{3 \times 3}$ represents respectively the position and the rotation matrix of the target with respect to the camera frame, $\tilde{p}_T^T \in \nR{4}$ indicates a homogeneous representation of target position with respect to camera frame, $u$ and $v$ are the coordinates of points projected in image 2D frame, and $f_x, f_y, c_x, c_y$ are camera known-parameters. Equation~(\ref{eq:pnp_maths}) establishes the correlation between 2D points observed by the camera and the corresponding 3D coordinates representing the target. Consequently, $\tilde{p}_T^T$ can be determined by executing the inverse operation in~(\ref{eq:pnp_maths}).

\subsection{Platform Description and Control Design} 
\label{sect:control}
The aerial manipulator platform can be modeled using decoupled dynamics for the \ac{UAV} and the manipulator. The decoupling simplifies the control strategy. The model of the platform is given by
\begin{subequations} \label{eq:sys_model}
\begin{align}
    m {\ddot{p}}(t) + m {G} &= F_p + F_{ext}, \label{eq:p_tau} \\
    \mathbf{J}(q(t)) {\ddot{q}}(t) + \mathbf{C}({q, \dot{q}, t}){\dot{q}} (t) &= {\tau_q} + \tau_{ext}, \label{eq:q_tau} \\
    \label{arm_model}
    \mathbf{M}_A \accA + \mathbf{D}_A \velA^A + \mathbf{K}_A \posA &= -F_A,
\end{align}    
\end{subequations}
where $p(t) \triangleq \begin{bmatrix}
    x(t), y(t), z(t)
\end{bmatrix}^T \in \mathbb{R}^3$, $q(t) \triangleq \begin{bmatrix}
    \phi(t), \theta(t), \psi(t)
\end{bmatrix}^T$ represents the position and orientation of the UAV in the inertial frame ($\mathbf{X_W - Y_W - Z_W}$); $G \triangleq \begin{bmatrix}
    0, 0, -g
\end{bmatrix}^T \in \mathbb{R}^3$ is the gravity vector with $g = 9.81 ms^{-2}$ is the acceleration due to gravity; $m, \mathbf{J}, \mathbf{C}$ are the mass inertia and Coriolis matrices of the UAV; $\tau_{q} \in \mathbb{R}^3$  is the attitude control input, $F_{ext}, \tau_{ext}$ are the external force and moment acting on the UAV from the manipulator, $\posA, \velA$ are the position and linear velocity of the manipulator respectively, $F_A$ represents the reaction forces on the end-effector, and $\mathbf{M}_A, \mathbf{D}_A, \mathbf{K}_A$ represent the inertia, damping and elasticity of mechanism. In this work, we consider the manipulator to be unactuated. The only external force acting on the end-effector will be due to the reaction forces from the physical interaction. For further details about the aerial manipulator design, please refer to \cite{incredible-arm}.


We further divide the control framework into position, velocity, and attitude control layers to simply the control design and introduce the safety functionality through \ac{CBF}. The CBF is used in the position controller to precisely establish contact with the target. Further, it prevents the UAV from deviating or colliding with the target forcefully. In this layer, a pseudo-velocity input is generated
\begin{align}
    u_W = \mathbf{K_p} (p_d - p),
\end{align}
where $\mathbf{K_p}$ is a positive definite gain matrix. This pseudo-velocity input is filtered using a CBF layer to produce the actual velocity input $u_W^*$. 

The CBF is designed to be a funnel-like function, which converges at the point of contact, as shown in Fig \ref{fig:frames}. Since velocity input is used in the position dynamics, the model is simplified to the following kinematic equation,

\begin{align}
	\dot{p} = u. \label{eq:kinematics}
\end{align}

The CBF function in this case is designed using a parabolic function, given by
\begin{align}
    h = x_D^T - a \sqrt{l},    
\end{align}
where $x_D^T, {y_D^T}, {z_D^T}$ are the location of the UAV with respect to the target frame, $l=\sqrt{{y_D^T}^2 + {z_D^T}^2}$ and $a$ is a design parameter for reshaping the CBF boundary. The derivatives of the CBF with respect to the states are given by
\begin{align}
    \frac{\partial h}{\partial x_D^T} &= 1, ~
    \frac{\partial h}{\partial y_D^T} = -\frac{a y_D^T}{2l^{3/4}}, ~ \frac{\partial h}{\partial z_D^T} = -\frac{a z_D^T}{2l^{3/4}}. \label{eq:partial_derivatives}
\end{align}

By the CBF definition in \cite{ames2019control}, the candidate CBF restricts the \ac{UAV} to operate only in the safe set and precisely establish contact with the target from any initial condition. The CBF constraint is used in the following quadratic equation to generate the velocity input, which ensures safety with only a minimal deviation from the pseudo velocity input,
\begin{align}
    u_T^* = \argmin_{u}||(u-u_T)|| \label{eq:quad_prog} \\
    \text{s.t. } \frac{\partial h}{\partial p_D^T}(f(x) + g(x)u) \geq -\omega(h) \label{eq:constraint}
\end{align}
where $\omega(h)$ is the optimization variable, $u_T = \mathbf{R}u_W$ is the nominal velocity input in target frame, $\mathbf{R}$ is the rotation matrix between world frame and target frame, $\frac{\partial h(\mathbf{p_D^T}, t)}{\partial t}=[\frac{\partial h}{\partial x_D^T} \;\;\frac{\partial h}{\partial y_D^T} \;\; \frac{\partial h}{\partial z_D^T}]^\top$ are as derived in Equation \eqref{eq:partial_derivatives} and $f(x) = 0, g(x) = \mathbf{I}$ from \eqref{eq:kinematics}. It is evident that for a given $p_T$, the constraint \eqref{eq:constraint} is linear in $u$, so that the quadratic equation can be solved at high speeds, using the CVXPY library from python for convex optimization problems. Since square roots are present in the constraint equation, a discontinuity occurs when $l = 0$, when the UAV is perfectly aligned with the target. However, the nominal controller can be used directly over this region. The obtained safe control input, $u_T*$ is converted to the world frame using the relationship, $u_W^* = \mathbf{R}^\top u_T^*$. Further, the yaw velocities are obtained as,
\begin{align}
    \dot{\psi}_d = K_{\psi}(\psi_d - \psi),
\end{align}
where $K_{\psi}$ is a positive scalar and $\psi_d$ is the desired yaw to establish the contact.

Now, the velocity tracking control and the lower level control are handled by the PX4's internal controller, to which we send the desired velocity inputs $u_W^*, \dot{\psi}_d$. Since the \ac{UAV} is an underactuated system, there is a partial coupling between the position and attitude dynamics. The desired attitude commands are chosen from the desired thrust along different axes. The internal controller uses a first PID controller to obtain the desired forces to track the input velocities. Then, it obtains the upward thrust and desired attitude from these forces. Further, it uses two PID controllers to track the attitude and angular velocities.

\section{System Description} 
\label{sect:sys_descr}
This section is devoted to presenting the various subsystems crucial for the proper evaluation of the methodology outlined in Section \ref{sect:method}. To validate the proposed approach, an aerial manipulator platform is presented (Section \ref{sect:platform}). The platform integrates an appropriate sensor setup for navigation and autonomous flight and a robotic arm for safe interaction with the environment. Another requirement for the validation of the approach is represented by the integration of target localization into the software architecture. The proposed solution for detecting target pose relies indeed on a Neural Network-based algorithm, which makes the integration with on-board computers not suitable for the proper performance of the Neural Network. To efficiently tackle this challenge, we have implemented an edge computing environment, which allows us to offboard the computation of \textit{target detection} to a \ac{k8s} clusters, as detailed in Section \ref{sect:edge}. Furthermore, Section \ref{sect:data} offers thorough insights into the communication aspects of the developed architecture, including communication within various subsystems onboard the drone and between the drone and \textit{target detection} executed in the cluster.

\subsection{Platform} 
\label{sect:platform}
For evaluating the proposed approach, a Tarot 650 served as the experimental platform. The drone is equipped with sensors to enable localization and secure interaction with the environment. Specifically, for target localization and detection, an \textit{Intel RealSense D455} depth camera is integrated into the platform. This camera utilizes color and depth images, that are required for marker detection, as detailed in Section \ref{sect:data}. Moreover, for precise target navigation, an \textit{Intel RealSense T265} tracking camera and a \textit{TF-mini} 1D \ac{LiDaR} are incorporated repeatedly for pose and depth measurements. Additionally, for flight controller testing, a Pixhawk with the PX4 Flight stack was integrated. The comprehensive list of sensors is provided in Table \ref{tab:sensor_setup}. 
To ensure safe interactions with contact surfaces when the target is reached, the flexible compliant arm introduced in \cite{incredible-arm} is integrated into the multirotor which enables the drone to perform physical interactions with the environment safely, without compromising its performance, due to its flexible properties and overall small weight. 

\begin{table}[ht!]
    \centering
    \begin{tabular}{|l|l|ll}
    \cline{1-2}
    \multicolumn{2}{|c|}{\textbf{Sensor Setup}}                                    &  &  \\ \cline{1-2}
    \multicolumn{1}{|c|}{\textbf{Components}} & \multicolumn{1}{|c|}{\textbf{Model}} &  &  \\ \cline{1-2}
    \multicolumn{1}{|l|}{IMU}                 & ICM-20948, ICM-20602 IMU         &  &  \\ \cline{1-2}                       
    \multicolumn{1}{|l|}{Color and Depth camera}        & Realsense D455                     &  &  \\ \cline{1-2}
    \multicolumn{1}{|l|}{Tracking Camera}     & Realsense T265                     &  &  \\ \cline{1-2}
    \multicolumn{1}{|l|}{1D \ac{LiDaR}}     & TF Mini                     &  &  \\ \cline{1-2}
    \end{tabular}
    \caption{Sensor setup integrated into Tarot T650 for experiments.}
    \label{tab:sensor_setup}
    \vspace{-0.5cm}
\end{table}

\subsection{Communication and Edge Offload} 
\label{sec:comm}
An inherent advantage of employing aerial robots for inspection lies in their ability to reduce costs and mitigate risks for human operators~\cite{9213854, 8967577, mansouri2017cooperative}. To further enhance cost-effectiveness, more affordable onboard computers are utilized. While many processes are executed on the onboard computer, the computational intensity of certain tasks necessitates offloading. Critical operations such as localization are executed on the onboard computer due to their time-sensitive nature. On the other hand, for tasks like target detection, which is crucial for successful inspections, an edge computing paradigm is employed to achieve optimal performance without introducing time delays~\cite{9740044, cheng2023real, alam2019uav}. This approach allows the execution of the resource-intensive neural network for target detection on a powerful edge computing system, leveraging heavy \ac{GPU}, preventing potential crashes on the onboard computer.

\subsubsection{Data Transmission} 
\label{sect:data}
In the proposed approach, data transmission for successful mission execution is divided into two layers (see Fig.~\ref{fig:offloading}). The first layer involves communication among \ac{ROS} nodes, both within the robot's onboard computer and in the edge computing environment. The second layer, based on the principles outlined in~\cite{damigos2024communication, damigos2023resilient}, utilizes a UDP tunnel for transmitting \ac{ROS} messages as UDP packets. This transmission occurs via Wi-Fi or cellular networks, facilitating bidirectional communication from the robot to the edge and back. This method effectively addresses the inherent challenges of \ac{ROS} communication between the internal cluster and the external cluster nodes.

\subsubsection{Edge Offloading} 
\label{sect:edge}
The proposed approach utilizes an edge computing environment, leveraging a \ac{k8s} cluster, to offload target detection algorithms. As illustrated in Fig.~\ref{fig:offloading}, the cluster processes camera data ($x_i(k) \in \mathbb{Z}^{n \times m}$, where $n \times m$ represents the image matrix size, at time step k) received from the robot and generating target localization ($u_i(k) \in \mathbb{R}^{1 \times 3}$) relative to the global frame. The system, however, encounters uplink ($d_1 \in \mathbb{R}^+$) and downlink ($d_2 \in \mathbb{R}^+$) delays. A filtering method is introduced to ensure accurate target localization; it disregards the localization output when the \ac{RRT} for the closed-loop system exceeds the maximum threshold $\tau^{max}$. Then, the operation of the system can be expressed as
\begin{align}
\label{eq:offloading}
    u_i(k) = \left\{
               \begin{array}{ll}
    f(x_i(k+d_1)), & \mbox{if} \hspace{2mm} d_1 + d_2 \leq \tau^{max} \\
    \mbox{``ignore"}, & \mbox{otherwise}
    \end{array}
    \right.
\end{align}
where $f(\cdot)$ is the target detection function executed in the edge computing environment.

The condition $d_1 + d_2 \leq \tau^{max}$ ensures that the target localization is only considered if the communication delay is within acceptable limits. If the delays exceed $\tau^{max}$, the target localization is disregarded to avoid using outdated or inaccurate data.

\begin{figure}[h]
    \centering
    \includegraphics[width=0.49\textwidth]{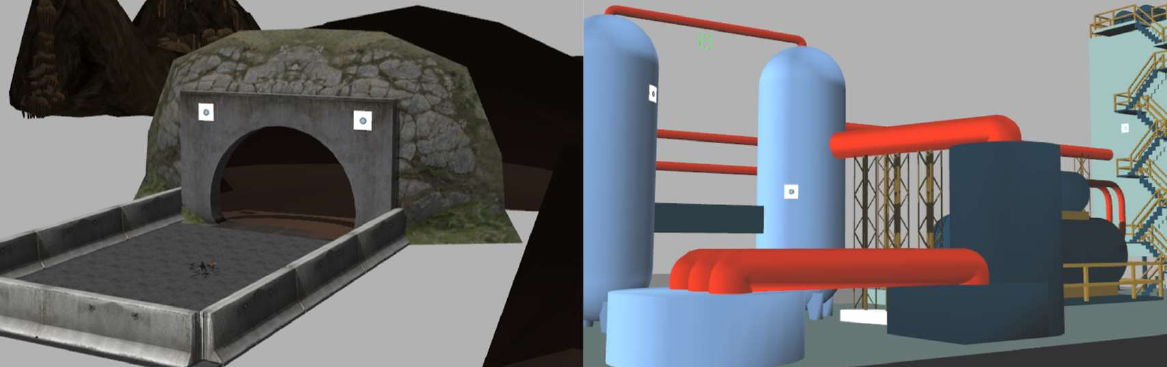}
    \caption{Simulation environments for the evaluation of the proposed architecture. Including tunnel (left) and refinery plant (right).}
    \label{fig:sim_enviroment}
    \vspace{-0.5cm}
\end{figure}

\begin{figure}[h] 
    \centering
    \begin{subfigure}{0.49\linewidth}
        \centering
        \includegraphics[width=\linewidth]{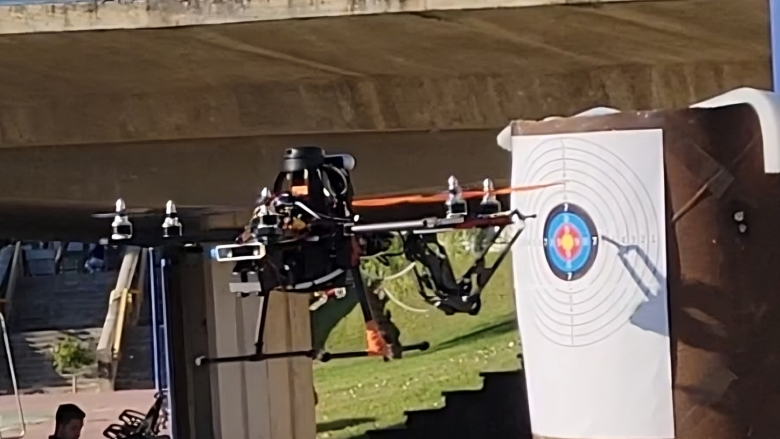}
        \caption{Real platform in industrial environment}
        \label{fig:platform_real}
    \end{subfigure}
    \begin{subfigure}{0.49\linewidth}
        \centering
        \includegraphics[width=\linewidth]{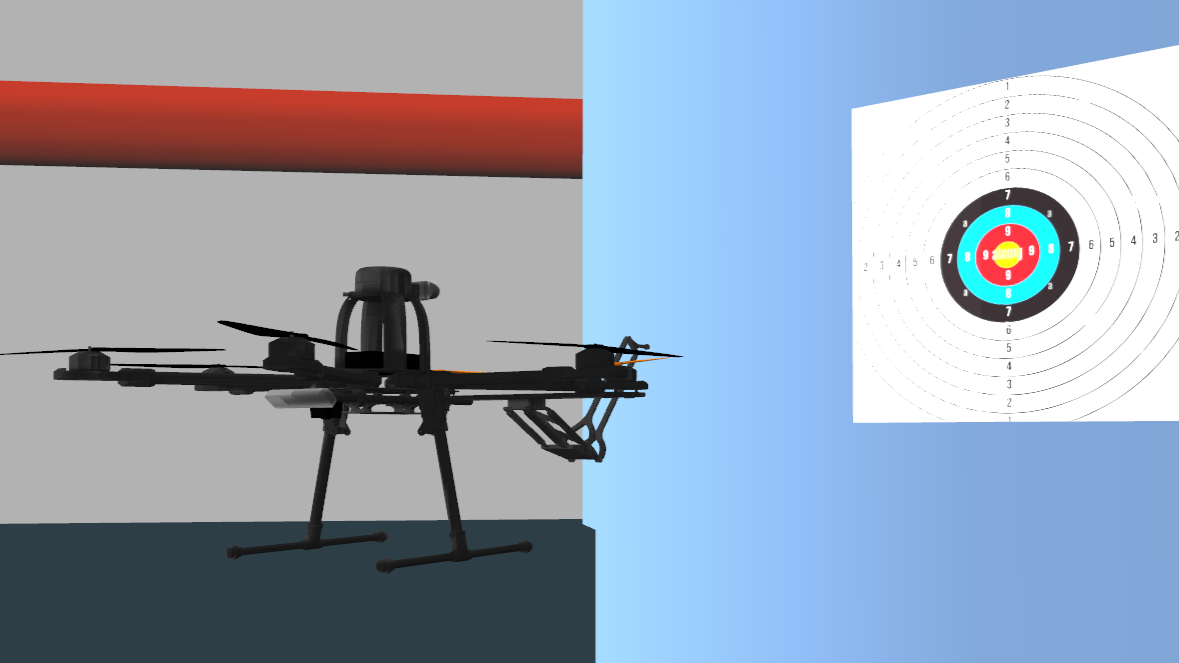}
        \caption{Platform in Gazebo simulation environment}
        \label{fig:platform_model}
    \end{subfigure}
    \caption{Side to side comparison of real and simulated \ac{UAV}.}
    \label{fig:real_vs_sim_env}
    \vspace{-0.5cm}
\end{figure}

\section{Experimental Applications} 
\label{sect:experiments}
The following section is devoted to the evaluation of the capabilities of the presented approach. In particular, Section \ref{sect:simulation} is devoted to the discussion of the results obtained in the Simulation environment, while Section \ref{sect:detect_evaluation} focuses on the experimental results obtained in real time.

\subsection{System Evaluation} 
\label{sect:simulation}
Along with the prototype, an advanced simulation environment has been designed, taking into account the aerial manipulator prototype, the sensor setup, and realistic world environments. The simulated version of the prototype is then incorporated into the Gazebo Simulator, following the procedure introduced in~\cite{airframe2023}, and into PX4 Software in the Loop (SITL) software for the testing phase (see Fig.~\ref{fig:sim_enviroment}). The simulated version includes a model of all the sensors integrated into the prototype (Table~\ref{tab:sensor_setup}) as well as an estimate of the mass and inertia of all components. A representation of the platform in simulation is presented in Fig.~\ref{fig:platform_model}. The development of the Simulator allows us to test the whole architecture in different simulation scenarios to evaluate the performance and robustness of the proposed solution. In particular, the experiments in the Simulation environment are performed in two different scenarios, introduced in~\cite{Markovic2023} and~\cite{koval2020subterranean}, to evaluate the behavior of the system under different light and environmental conditions (see Fig.~\ref{fig:real_vs_sim_env}). 

\begin{figure}
    \centering
    \includegraphics[width=0.45\textwidth]{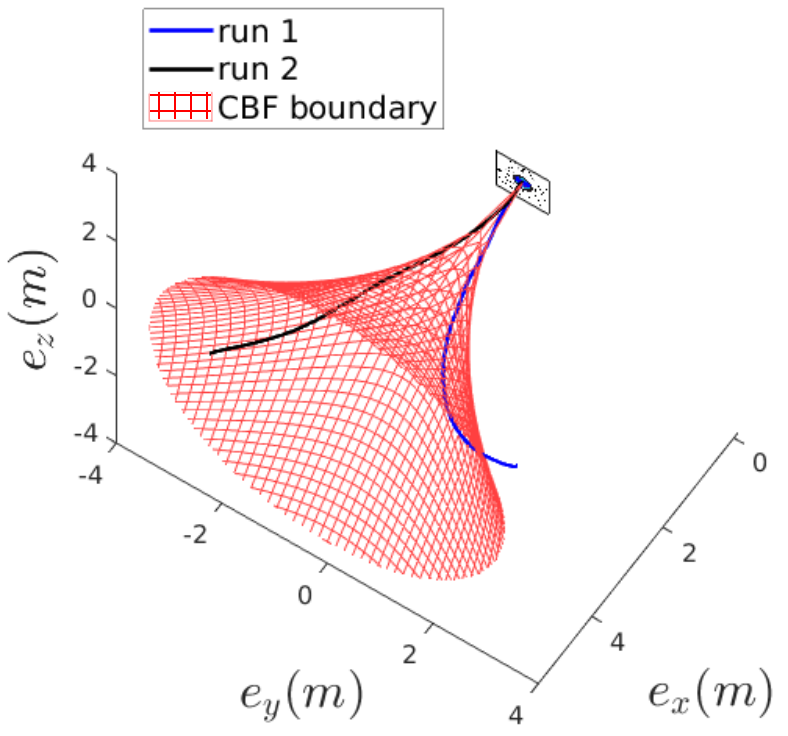}
    \caption{The CBF boundary layer ($a=3$) along with the trajectories of the UAV guided by the CBF to the target.}
    \label{fig:cbf_3d}
    \vspace{-0.5cm}
\end{figure}

\begin{figure}
    \centering
    \includegraphics[width=0.48\textwidth]{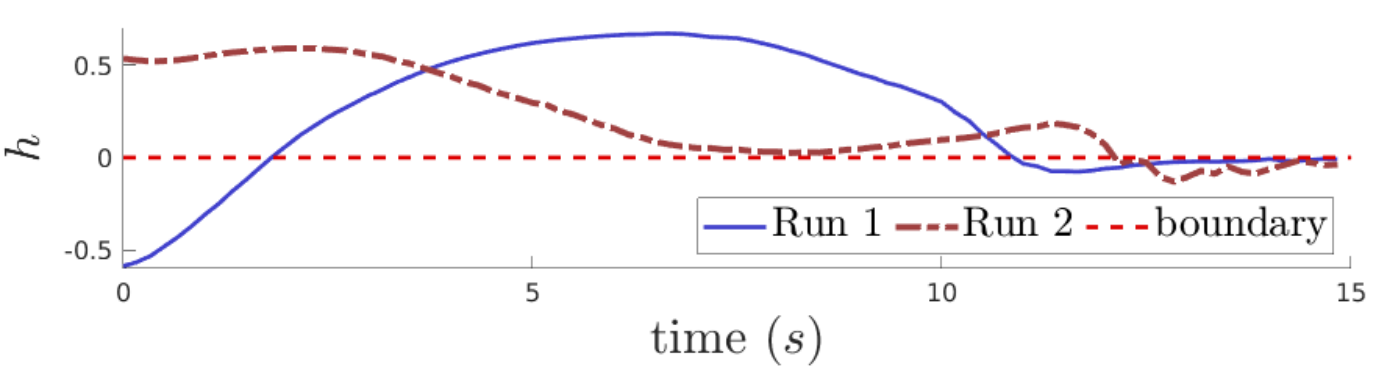}
    \caption{The values of $h$ during Run 1 and Run 2.}
    \label{fig:h_func}
    \vspace{-0.5cm}
\end{figure}

\begin{figure}
    \centering
    \includegraphics[width=0.48\textwidth]{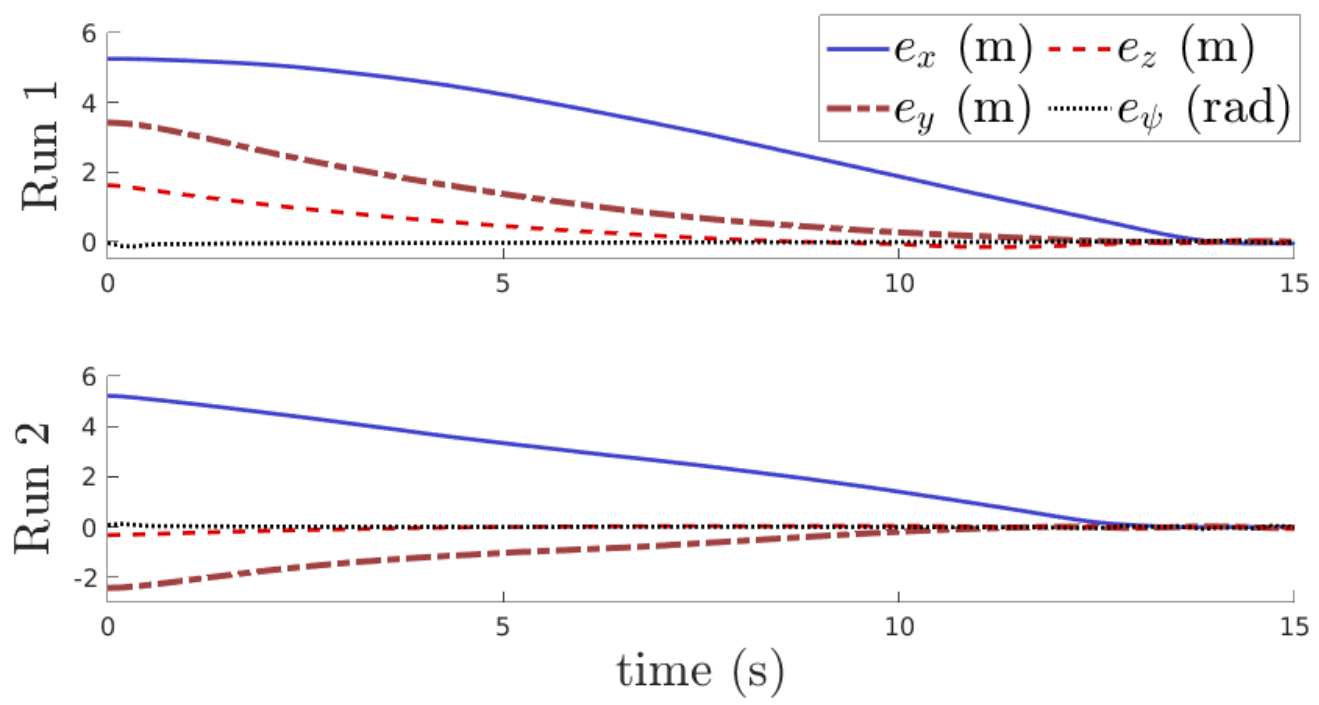}
    \caption{Errors in position and yaw of the UAV for the two runs.}
    \label{fig:pos_err}
    \vspace{-0.5cm}
\end{figure}

\begin{figure}
    \centering
    \includegraphics[width=0.48\textwidth]{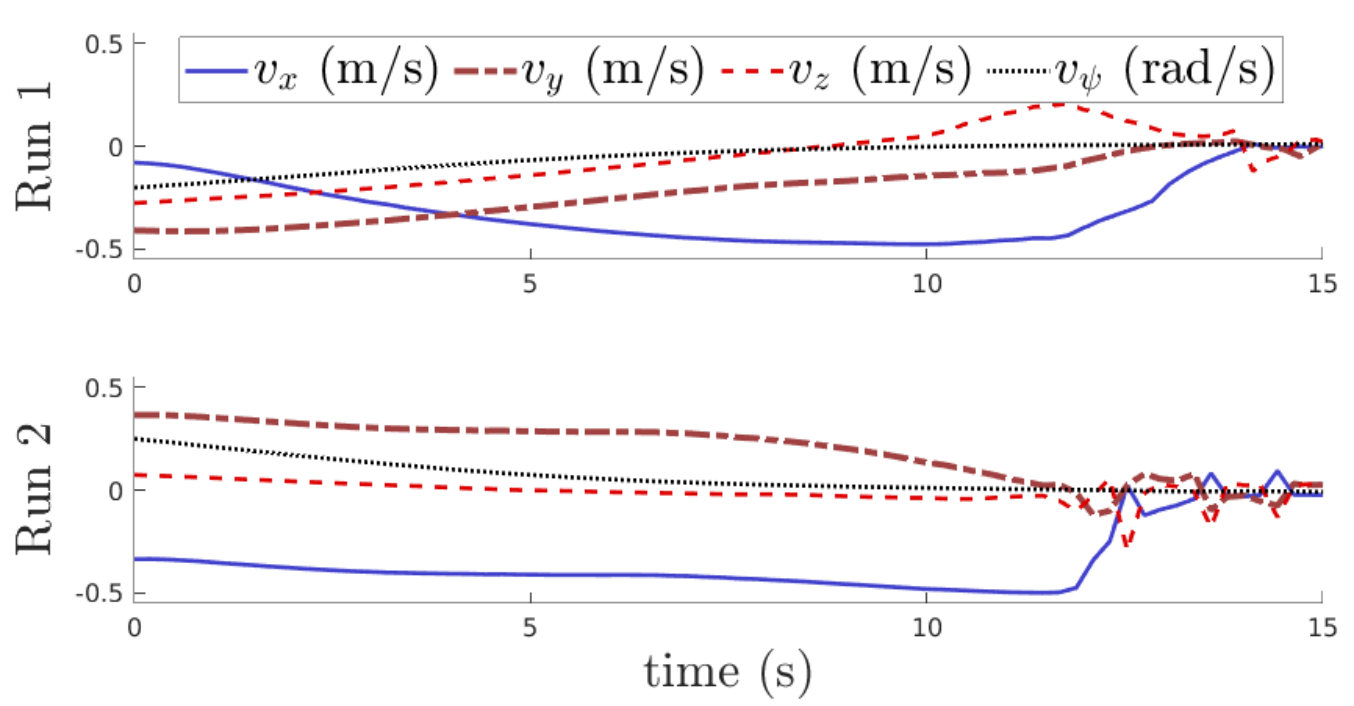}
    \caption{Commanded velocity for the UAV for the runs.}
    \label{fig:cmd_vel}
    \vspace{-0.5cm}
\end{figure}

The performance of the controller is validated in the simulated environments, which is highlighted in Fig. \ref{fig:cbf_3d} - \ref{fig:cmd_vel}. In the simulated scenario, the controller is commanded to precisely navigate the UAV to the given target's pose (obtained from the target detection framework). Fig. \ref{fig:cbf_3d} visualizes how the \ac{CBF} sets a boundary (red mesh) for the \ac{UAV} to maneuver safely towards the target without requiring an external planner, where run 1 corresponds to the cave environment, and run 2 corresponds to the refinery environment. It is to be noted that the mesh and the position error trajectories ($e_x, e_y, e_z$ in Fig. \ref{fig:cbf_3d}, \ref{fig:pos_err}) are the same as the position trajectories of the UAV in the target frame ($x_D^T, y_D^T, z_D^T$). In run 1, though the UAV starts from an unsafe region, the trajectory converges to the safe zone and approaches the target through it. The action of the CBF is further evident from Fig. \ref{fig:pos_err}, where the error trajectories $e_y, e_z$ converge to zero much faster than $e_x$, showing that the UAV aligns itself right in front of the target before approaching it and hence forming a parabolic trajectory as seen in Fig. \ref{fig:cbf_3d}. Further, in Fig. \ref{fig:h_func}, the asymptotic convergence to the safe zone, while starting from an unsafe zone is evident in Run 1, where the $h$ value starts from a negative value and rapidly increases to a positive value. When a positive value is reached, it approaches the target horizontally. After around 12 seconds, the values of $h$ in both trajectories get close to zero when the \ac{UAV} gets close to the target, where the safe region narrows into a point. It can be observed that there are steep changes in the commanded control inputs in Fig. \ref{fig:cmd_vel}, as the \ac{CBF} tries to push the \ac{UAV} to remain in the safe zone. However, due to the actuator's physical limitations, the \ac{UAV} gets into the unsafe zone momentarily. So, the $h$ values, and hence the commanded velocities slowly oscillate near the target to maintain the position error as close as possible to zero, which is observed in Fig. \ref{fig:pos_err}. When the \ac{UAV} reaches the tip of the funnel, it establishes contact with the target safely with a very low velocity.

\subsection{Target Detection Evaluation}
\label{sect:detect_evaluation}
\begin{figure}[t!] 
    \centering
    \begin{subfigure}{0.49\linewidth}
        \centering
        \includegraphics[width=\linewidth]{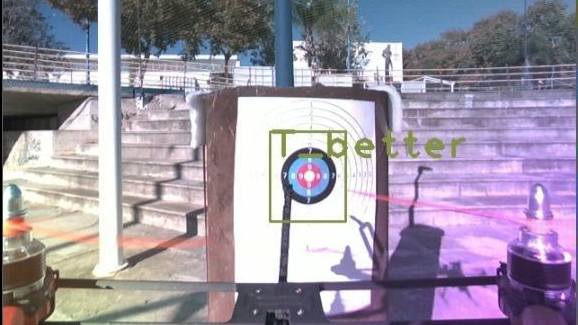}
        \caption{Color image outdoor}
        \label{fig:test_detection-seville-1}
    \end{subfigure}
    \begin{subfigure}{0.49\linewidth}
        \centering
        \includegraphics[width=\linewidth]{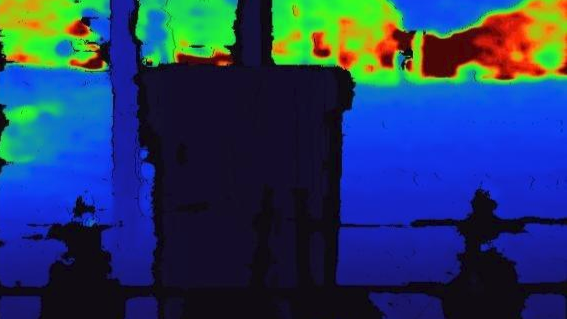}
        \caption{Depth image outdoor}
        \label{fig:test_detection-seville-2}
    \end{subfigure}
    \\
    \begin{subfigure}{0.49\linewidth}
        \centering
        \includegraphics[width=\linewidth]{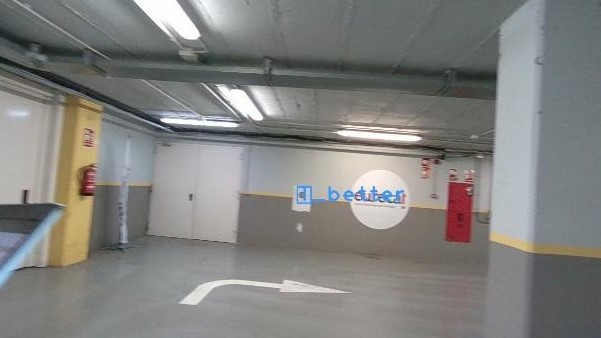}
        \caption{Color image in parking}
        \label{fig:test_detection-parking-1}
    \end{subfigure}
    \begin{subfigure}{0.49\linewidth}
        \centering
        \includegraphics[width=\linewidth]{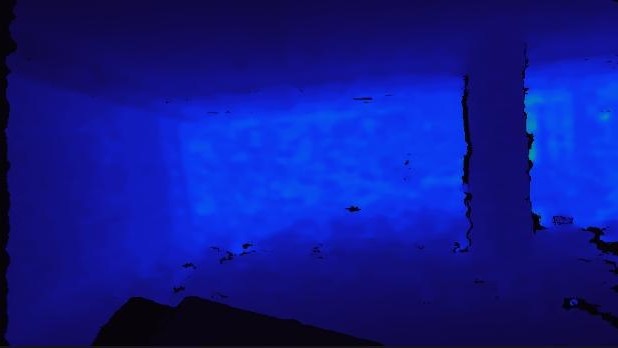}
        \caption{Depth image in parking}
        \label{fig:test_detection-parking-2}
    \end{subfigure}
    \\
    \begin{subfigure}{0.49\linewidth}
        \centering
        \includegraphics[width=\linewidth]{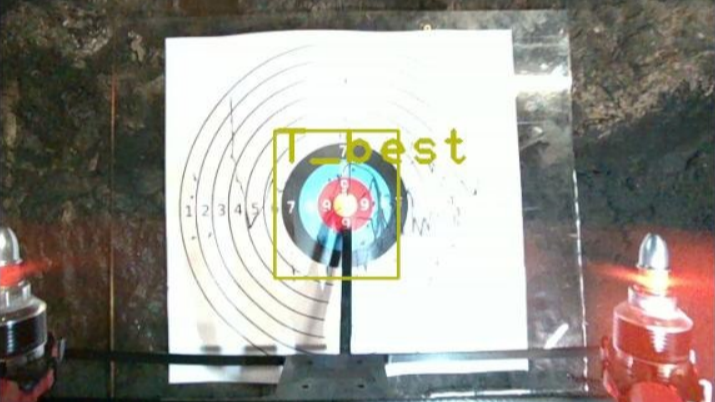}
        \caption{Color image in tunnel}
        \label{fig:test_detection-tunnel-1}
    \end{subfigure}
    \begin{subfigure}{0.49\linewidth}
        \centering
        \includegraphics[width=\linewidth]{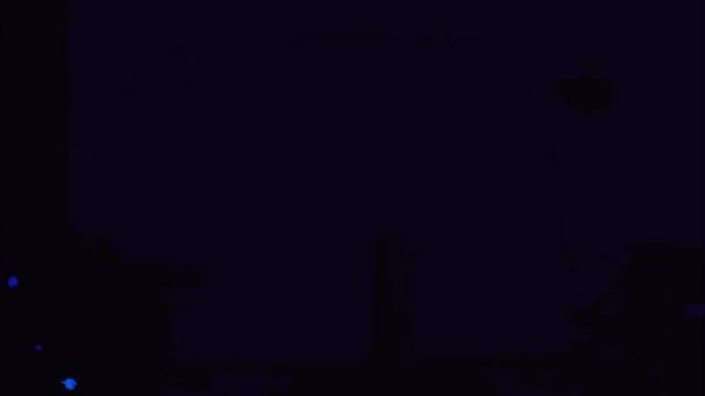}
        \caption{Depth image in tunnel}
        \label{fig:test_detection-tunnel-2}
    \end{subfigure}
    \\
    \begin{subfigure}{0.49\linewidth}
        \centering
        \includegraphics[width=\linewidth]{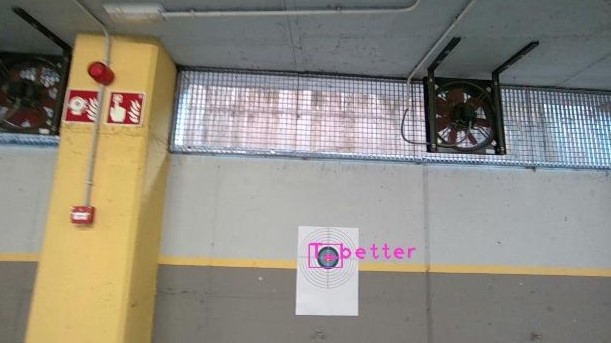}
        \caption{Rich background}
        \label{fig:test_detection-5}
    \end{subfigure}
    \begin{subfigure}{0.49\linewidth}
        \centering
        \includegraphics[width=\linewidth]{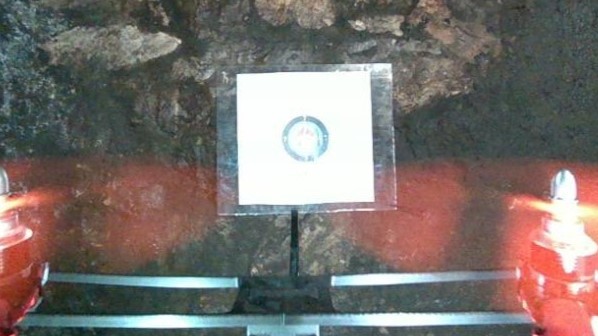}
        \caption{Reflection}
        \label{fig:test_detection-6}
    \end{subfigure}
    \caption{Color and depth images of target detection in real flight application with previously unseen images.}
    \label{fig:test_target_detection}
    \vspace{-0.5cm}
\end{figure}

The validation of the detection algorithm was conducted across various scenarios, each characterized by distinct lighting and environmental conditions, facilitating a comprehensive assessment of the proposed approach's performance and robustness. In particular, experiments were conducted in three different locations: a parking lot, a tunnel, and an outdoor environment during midday with intense sunlight. Each location presented unique challenges, allowing for a thorough evaluation of target detection performance. Visual testing was performed during each flight, and Fig.~\ref{fig:test_target_detection} illustrates three samples showcasing the detector's output alongside depth images. From the image, it's possible to notice the different light conditions that affect the color and depth: the first and second scenarios are particularly dark, as evidenced by the lack of discernible information in the depth images. Additionally, the second scenario posed additional challenges due to reflections generated by the target cover, Fig.~\ref{fig:test_detection-6}. In contrast, the third scenario presents contrasting conditions with ample sunlight, resulting in potential overexposure of the image sensor and a resulting pink hue in the images.

\begin{figure}[h!]
    \centering
    \includegraphics[width=0.48\textwidth]{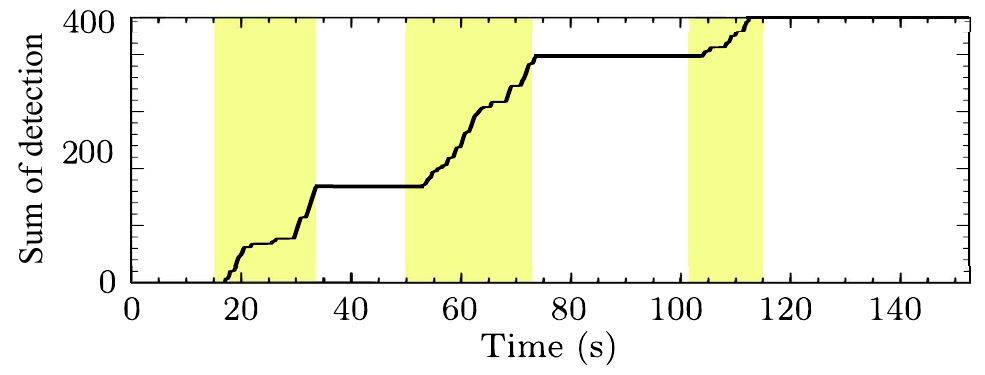}
    \caption{Plot of the sum of target detection through time during a flight test. Regions in yellow highlight moments when a target is within the camera's field of view.}
    \label{fig:target_detection_perf}
\end{figure}

Despite the different conditions faced, the detector consistently demonstrated robust performance, achieving $T\_best$ performance label detection even with a scratched target (see Fig.~\ref{fig:test_detection-tunnel-1}), and $T\_better$ in long-distance estimation (see Fig.~\ref{fig:test_detection-parking-1}). This last detection is also impressive due to the proximity of a spherical pattern that could have been detected as a false positive. Fig.~\ref{fig:test_detection-5} also shows the detection quality in a rich features environment. Nevertheless, it is interesting to notice that light reflections on the target surface in the tunnel lead to false negative detection.

Due to the onboard computer's inability to process data, execute the detection algorithm, and determine the target's location in real time, we employed the edge offloading strategy. Camera data were streamed to the \ac{k8s} cluster for target detection and localization. Subsequently, the target's location was transmitted to the aerial platform for its navigation.

To further evaluate the algorithm's performance, we conducted additional experiments in the tunnel, which is the most demanding environment. The test consists of a flight with multiple fly-bys to different targets attached to the walls. We evaluate the number of positive target detections during the all flight. The result is presented in Figure \ref{fig:target_detection_perf}. It shows that whenever a target enters the field of view, there is an increase in the number of target detections. Conversely, this count remains constant when the target exits the camera's field of view, showing consistency of true negatives. \\
Overall, the behavior of the algorithm and the quality of detection exhibit robustness and consistency across various flight tests and scenarios. The algorithm consistently demonstrates strong performance, successfully detecting targets with reliability regardless of the environmental conditions or flight scenarios encountered.

\section{Conclusions and Future Work} 
\label{sect:conclusions}

In this work, we introduced a refined approach to autonomously and safely engage with a target, blending neural networks with traditional control methods. By leveraging deep learning solely for image processing and maintaining a model-based planner and controller, we ensured precise control outputs. Utilizing edge computing, we offloaded the computationally intensive neural tasks requiring a \ac{GPU}, preserving the lightweight nature of the \ac{MAV} while ensuring mission oversight. Furthermore, employing an onboard CBF-based controller enables the \ac{UAV} to navigate safely and accurately establish contact with the target without relying on an external planner. Experimental validation of the system underscored the effectiveness of the proposed architecture.

Future steps will include the evaluation of the whole architecture in the real world and the integration of the exploration phase, to enable \ac{UAV} to autonomously conduct scheduled, and safe \ac{APhI} tasks in industrial environments.

\bibliographystyle{IEEEtran}
\bibliography{references}

\begin{thebibliography}{10}
\providecommand{\url}[1]{#1}
\csname url@samestyle\endcsname
\providecommand{\newblock}{\relax}
\providecommand{\bibinfo}[2]{#2}
\providecommand{\BIBentrySTDinterwordspacing}{\spaceskip=0pt\relax}
\providecommand{\BIBentryALTinterwordstretchfactor}{4}
\providecommand{\BIBentryALTinterwordspacing}{\spaceskip=\fontdimen2\font plus
\BIBentryALTinterwordstretchfactor\fontdimen3\font minus \fontdimen4\font\relax}
\providecommand{\BIBforeignlanguage}[2]{{%
\expandafter\ifx\csname l@#1\endcsname\relax
\typeout{** WARNING: IEEEtran.bst: No hyphenation pattern has been}%
\typeout{** loaded for the language `#1'. Using the pattern for}%
\typeout{** the default language instead.}%
\else
\language=\csname l@#1\endcsname
\fi
#2}}
\providecommand{\BIBdecl}{\relax}
\BIBdecl

\bibitem{ollero2019aerial}
A.~Ollero and B.~Siciliano, \emph{Aerial robotic manipulation}.\hskip 1em plus 0.5em minus 0.4em\relax Springer, 2019.

\bibitem{nikolakopoulos2022aerial}
G.~Nikolakopoulos, S.~S. Mansouri, and C.~Kanellakis, \emph{Aerial Robotic Workers: Design, Modeling, Control, Vision and Their Applications}.\hskip 1em plus 0.5em minus 0.4em\relax Butterworth-Heinemann, 2022.

\bibitem{past-future-am}
A.~Ollero, M.~Tognon, A.~Suarez, D.~Lee, and A.~Franchi, ``Past, present, and future of aerial robotic manipulators,'' \emph{IEEE Transactions on Robotics}, vol.~38, no.~1, pp. 626--645, 2022.

\bibitem{bridge-inspection}
A.~Viguria, R.~Caballero, {\'A}.~Petrus, F.~J. P{\'e}rez-Grau, and M.~{\'A}. Trujillo, ``Aerial robotic system for complete bridge inspections,'' in \emph{Advances in Road Infrastructure and Mobility}.\hskip 1em plus 0.5em minus 0.4em\relax Cham: Springer International Publishing, 2022, pp. 171--186.

\bibitem{zhu2022rapid}
Y.~Zhu, J.~Zhou, Y.~Yang, L.~Liu, F.~Liu, and W.~Kong, ``Rapid target detection of fruit trees using uav imaging and improved light yolov4 algorithm,'' \emph{Remote Sensing}, vol.~14, no.~17, p. 4324, 2022.

\bibitem{drones6110335}
\BIBentryALTinterwordspacing
Y.~Li, H.~Yuan, Y.~Wang, and C.~Xiao, ``Ggt-yolo: A novel object detection algorithm for drone-based maritime cruising,'' \emph{Drones}, vol.~6, no.~11, 2022. [Online]. Available: \url{https://www.mdpi.com/2504-446X/6/11/335}
\BIBentrySTDinterwordspacing

\bibitem{f14091812}
\BIBentryALTinterwordspacing
Z.~Xiao, F.~Wan, G.~Lei, Y.~Xiong, L.~Xu, Z.~Ye, W.~Liu, W.~Zhou, and C.~Xu, ``Fl-yolov7: A lightweight small object detection algorithm in forest fire detection,'' \emph{Forests}, vol.~14, no.~9, 2023. [Online]. Available: \url{https://www.mdpi.com/1999-4907/14/9/1812}
\BIBentrySTDinterwordspacing

\bibitem{s16111778}
\BIBentryALTinterwordspacing
J.~Sun, B.~Li, Y.~Jiang, and C.-y. Wen, ``A camera-based target detection and positioning uav system for search and rescue (sar) purposes,'' \emph{Sensors}, vol.~16, no.~11, 2016. [Online]. Available: \url{https://www.mdpi.com/1424-8220/16/11/1778}
\BIBentrySTDinterwordspacing

\bibitem{jiang2022review}
P.~Jiang, D.~Ergu, F.~Liu, Y.~Cai, and B.~Ma, ``A review of yolo algorithm developments,'' \emph{Procedia Computer Science}, vol. 199, pp. 1066--1073, 2022.

\bibitem{kou2020research}
M.~Kou, L.~Zhou, J.~Zhang, and H.~Zhang, ``Research advances on object detection in unmanned aerial vehicle imagery,'' \emph{Meas. Control. Technol}, vol.~39, pp. 47--61, 2020.

\bibitem{TAN2021107261}
\BIBentryALTinterwordspacing
L.~Tan, X.~Lv, X.~Lian, and G.~Wang, ``Yolov4\_drone: Uav image target detection based on an improved yolov4 algorithm,'' \emph{Computers \& Electrical Engineering}, vol.~93, p. 107261, 2021. [Online]. Available: \url{https://www.sciencedirect.com/science/article/pii/S0045790621002445}
\BIBentrySTDinterwordspacing

\bibitem{9372841}
P.~Pradeepkumar, A.~Pal, and K.~Kant, ``Resource efficient edge computing infrastructure for video surveillance,'' \emph{IEEE Transactions on Sustainable Computing}, pp. 1--1, 2021.

\bibitem{8479091}
C.~O. Bitye~Dimithe, C.~Reid, and B.~Samata, ``Offboard machine learning through edge computing for robotic applications,'' in \emph{SoutheastCon 2018}, 2018, pp. 1--7.

\bibitem{ganguly2021efficient}
S.~Ganguly, V.~N. Sankaranarayanan, B.~Suraj, R.~D. Yadav, and S.~Roy, ``Efficient manoeuvring of quadrotor under constrained space and predefined accuracy,'' in \emph{2021 IEEE/RSJ International Conference on Intelligent Robots and Systems (IROS)}.\hskip 1em plus 0.5em minus 0.4em\relax IEEE, pp. 6352--6357.

\bibitem{sankaranarayanan2022robustifying}
V.~N. Sankaranarayanan, R.~D. Yadav, R.~K. Swayampakula, S.~Ganguly, and S.~Roy, ``Robustifying payload carrying operations for quadrotors under time-varying state constraints and uncertainty,'' \emph{IEEE Robotics and Automation Letters}, vol.~7, no.~2, pp. 4885--4892, 2022.

\bibitem{sankaranarayanan2023adaptive}
V.~N. Sankaranarayanan, S.~Satpute, S.~Roy, and G.~Nikolakopoulos, ``Adaptive control of euler-lagrange systems under time-varying state constraints without a priori bounded uncertainty,'' \emph{IFAC-PapersOnLine}, vol.~56, no.~2, pp. 3360--3365, 2023.

\bibitem{ames2019control}
A.~D. Ames, S.~Coogan, M.~Egerstedt, G.~Notomista, K.~Sreenath, and P.~Tabuada, ``Control barrier functions: Theory and applications,'' in \emph{2019 18th European control conference (ECC)}.\hskip 1em plus 0.5em minus 0.4em\relax IEEE, 2019, pp. 3420--3431.

\bibitem{SAVIOLO202345}
\BIBentryALTinterwordspacing
A.~Saviolo and G.~Loianno, ``Learning quadrotor dynamics for precise, safe, and agile flight control,'' \emph{Annual Reviews in Control}, vol.~55, pp. 45--60, 2023. [Online]. Available: \url{https://www.sciencedirect.com/science/article/pii/S1367578823000135}
\BIBentrySTDinterwordspacing

\bibitem{aerial_robotics_springer}
A.~Ollero and B.~Siciliano, \emph{Aerial Robotic Manipulation}.\hskip 1em plus 0.5em minus 0.4em\relax Springer, 2019.

\bibitem{o2020deep}
N.~O’Mahony, S.~Campbell, A.~Carvalho, S.~Harapanahalli, G.~V. Hernandez, L.~Krpalkova, D.~Riordan, and J.~Walsh, ``Deep learning vs. traditional computer vision,'' in \emph{Advances in Computer Vision: Proceedings of the 2019 Computer Vision Conference (CVC), Volume 1 1}.\hskip 1em plus 0.5em minus 0.4em\relax Springer, 2020, pp. 128--144.

\bibitem{qing2021collision}
W.~Qing, H.~Chen, X.~Wang, and Y.~Yin, ``Collision-free trajectory generation for uav swarm formation rendezvous,'' in \emph{2021 IEEE International Conference on Robotics and Biomimetics (ROBIO)}.\hskip 1em plus 0.5em minus 0.4em\relax IEEE, 2021, pp. 1861--1867.

\bibitem{lee2016vision}
H.~Lee, S.~Jung, and D.~H. Shim, ``Vision-based uav landing on the moving vehicle,'' in \emph{2016 International conference on unmanned aircraft systems (ICUAS)}.\hskip 1em plus 0.5em minus 0.4em\relax IEEE, 2016, pp. 1--7.

\bibitem{lippi2021control}
M.~Lippi and A.~Marino, ``A control barrier function approach to human-multi-robot safe interaction,'' in \emph{2021 29th Mediterranean Conference on Control and Automation (MED)}.\hskip 1em plus 0.5em minus 0.4em\relax IEEE, 2021, pp. 604--609.

\bibitem{ferraguti2020control}
F.~Ferraguti, M.~Bertuletti, C.~T. Landi, M.~Bonf{\`e}, C.~Fantuzzi, and C.~Secchi, ``A control barrier function approach for maximizing performance while fulfilling to iso/ts 15066 regulations,'' \emph{IEEE Robotics and Automation Letters}, vol.~5, no.~4, pp. 5921--5928, 2020.

\bibitem{liang2023adaptive}
J.~Liang, H.~Zhong, Y.~Wang, Y.~Chen, J.~Zeng, and J.~Mao, ``Adaptive force tracking impedance control for aerial interaction in uncertain contact environment using barrier function,'' \emph{IEEE Transactions on Automation Science and Engineering}, 2023.

\bibitem{lin2023gnss}
H.-Y. Lin and J.-R. Zhan, ``Gnss-denied uav indoor navigation with uwb incorporated visual inertial odometry,'' \emph{Measurement}, vol. 206, p. 112256, 2023.

\bibitem{10275007}
H.~Luo, G.~Li, D.~Zou, K.~Li, X.~Li, and Z.~Yang, ``Uav navigation with monocular visual inertial odometry under gnss-denied environment,'' \emph{IEEE Transactions on Geoscience and Remote Sensing}, vol.~61, pp. 1--15, 2023.

\bibitem{unknown-author-no-date}
\BIBentryALTinterwordspacing
``{Visual Intertial Odometry (VIO) | PX4 User Guide (v1.12)}.'' [Online]. Available: \url{https://docs.px4.io/v1.12/en/computer\_vision/}
\BIBentrySTDinterwordspacing

\bibitem{5184844}
M.~Fiala, ``Designing highly reliable fiducial markers,'' \emph{IEEE Transactions on Pattern Analysis and Machine Intelligence}, vol.~32, no.~7, pp. 1317--1324, 2010.

\bibitem{kentaro_wada_2021_5711226}
\BIBentryALTinterwordspacing
K.~Wada, ``wkentaro/labelme,'' Nov. 2021. [Online]. Available: \url{https://doi.org/10.5281/zenodo.5711226}
\BIBentrySTDinterwordspacing

\bibitem{marchand2015pose}
E.~Marchand, H.~Uchiyama, and F.~Spindler, ``Pose estimation for augmented reality: a hands-on survey,'' \emph{IEEE transactions on visualization and computer graphics}, vol.~22, no.~12, pp. 2633--2651, 2015.

\bibitem{incredible-arm}
J.~Mellet, A.~Berra, A.~Santi~Seisa, V.~Sankaranarayanan, U.~Gamage, M.~Trujillo, G.~Heredia, G.~Nikolakopoulos, V.~Lippiello, and F.~Ruggiero, ``Design of a flexible robot arm for safe aerial physical interaction,'' in \emph{2024 IEEE 7th International Conference on Soft Robotics (RoboSoft)}.\hskip 1em plus 0.5em minus 0.4em\relax IEEE, 2024.

\bibitem{9213854}
M.~Bobbe, Y.~Khedar, J.~Backhaus, M.~Gerke, Y.~Ghassoun, and F.~Plöger, ``Reactive mission planning for uav based crane rail inspection in an automated container terminal,'' in \emph{2020 International Conference on Unmanned Aircraft Systems (ICUAS)}, 2020, pp. 1286--1293.

\bibitem{8967577}
R.~S. Pahwa, K.~Y. Chan, J.~Bai, V.~B. Saputra, M.~N. Do, and S.~Foong, ``Dense 3d reconstruction for visual tunnel inspection using unmanned aerial vehicle,'' in \emph{2019 IEEE/RSJ International Conference on Intelligent Robots and Systems (IROS)}, 2019, pp. 7025--7032.

\bibitem{mansouri2017cooperative}
S.~S. Mansouri, C.~Kanellakis, E.~Fresk, D.~Kominiak, and G.~Nikolakopoulos, ``Cooperative uavs as a tool for aerial inspection of the aging infrastructure,'' in \emph{Field and Service Robotics: Results of the 11th International Conference}.\hskip 1em plus 0.5em minus 0.4em\relax Springer, 2017, pp. 177--189.

\bibitem{9740044}
S.~Liang, H.~Wu, L.~Zhen, Q.~Hua, S.~Garg, G.~Kaddoum, M.~M. Hassan, and K.~Yu, ``Edge yolo: Real-time intelligent object detection system based on edge-cloud cooperation in autonomous vehicles,'' \emph{IEEE Transactions on Intelligent Transportation Systems}, vol.~23, no.~12, pp. 25\,345--25\,360, 2022.

\bibitem{cheng2023real}
Q.~Cheng, H.~Wang, B.~Zhu, Y.~Shi, and B.~Xie, ``A real-time uav target detection algorithm based on edge computing,'' \emph{Drones}, vol.~7, no.~2, p.~95, 2023.

\bibitem{alam2019uav}
M.~S. Alam, B.~Natesha, T.~Ashwin, and R.~M.~R. Guddeti, ``Uav based cost-effective real-time abnormal event detection using edge computing,'' \emph{Multimedia tools and Applications}, vol.~78, pp. 35\,119--35\,134, 2019.

\bibitem{damigos2024communication}
G.~Damigos, N.~Stathoulopoulos, A.~Koval, T.~Lindgren, and G.~Nikolakopoulos, ``Communication-aware control of large data transmissions via centralized cognition and 5g networks for multi-robot map merging,'' \emph{Journal of Intelligent \& Robotic Systems}, vol. 110, no.~1, p.~22, 2024.

\bibitem{damigos2023resilient}
G.~Damigos, A.~S. Seisa, S.~G. Satpute, T.~Lindgren, and G.~Nikolakopoulos, ``A resilient framework for 5g-edge-connected uavs based on switching edge-mpc and onboard-pid control,'' in \emph{2023 IEEE 32nd International Symposium on Industrial Electronics (ISIE)}.\hskip 1em plus 0.5em minus 0.4em\relax IEEE, 2023, pp. 1--8.

\bibitem{airframe2023}
A.~Berra, P.~J. Sanchez-Cuevas, M.~Trujillo, G.~Heredia, and A.~Viguria, ``Airframe - fast prototyping framework for uavs definition,'' in \emph{2023 International Conference on Unmanned Aircraft Systems (ICUAS)}, 2023, pp. 1175--1182.

\bibitem{Markovic2023}
\BIBentryALTinterwordspacing
L.~Markovic, F.~Petric, A.~Ivanovic, J.~Goricanec, M.~Car, M.~Orsag, and S.~Bogdan, ``Towards a standardized aerial platform: {ICUAS}'22 firefighting competition,'' \emph{Journal of Intelligent Robotic Systems}, vol. 108, no.~3, Jul. 2023. [Online]. Available: \url{https://doi.org/10.1007/s10846-023-01909-z}
\BIBentrySTDinterwordspacing

\bibitem{koval2020subterranean}
A.~Koval, C.~Kanellakis, E.~Vidmark, J.~Haluska, and G.~Nikolakopoulos, ``A subterranean virtual cave world for gazebo based on the darpa subt challenge,'' \emph{arXiv preprint arXiv:2004.08452}, 2020.

\end{thebibliography}

\end{document}